\documentclass[10pt,twocolumn,letterpaper]{article}

\usepackage[pagenumbers]{cvpr} 

\usepackage{algorithm}
\usepackage{algpseudocode}   
\usepackage{setspace}

\usepackage{wrapfig}
\usepackage{amssymb}
\usepackage{pifont}
\usepackage{bbding}
\usepackage{makecell}

\usepackage{adjustbox}
\usepackage{multirow}
\usepackage{graphicx}
\usepackage{xfrac}
\usepackage{arydshln}
\usepackage{adjustbox}
\usepackage{subcaption}
\usepackage{multirow}
\usepackage{booktabs}
\usepackage{siunitx}
\usepackage{xcolor}
\usepackage{listings}

\usepackage[T1]{fontenc}
\usepackage{textcomp}

\renewcommand{\paragraph}[1]{\vspace{.4em}\noindent\textbf{#1}}

\definecolor{codecomment}{rgb}{0.25,0.5,0.5}
\definecolor{codekeyword}{rgb}{0.35,0.35,0.75}
\definecolor{codestring}{rgb}{0.60,0.20,0.20}

\lstdefinestyle{py}{
  language=Python,
  basicstyle=\ttfamily\fontsize{7.2pt}{7.2pt}\selectfont,
  columns=fullflexible,
  breaklines=true,
  breakatwhitespace=true,
  showstringspaces=false,
  upquote=true,
  commentstyle=\color{codecomment},
  keywordstyle=\color{codekeyword}\bfseries,
  stringstyle=\color{codestring},
  tabsize=4,
  keepspaces=true,
  aboveskip=2pt,
  belowskip=2pt,
}
\usepackage{amssymb}
\usepackage{amsmath,amsfonts}
\usepackage{amsopn}
\usepackage{bm} %
\usepackage{multirow}
\usepackage{tabularx}

\newcommand{\vct}[1]{\boldsymbol{#1}} %

\newcommand{\ProbOpr}[1]{\mathbb{#1}}

\newcommand{\expect}[2]{%
\ifthenelse{\equal{#2}{}}{\ProbOpr{E}_{#1}}
{\ifthenelse{\equal{#1}{}}{\ProbOpr{E}\left[#2\right]}{\ProbOpr{E}_{#1}\left[#2\right]}}} %
\newcommand{\var}[2]{%
\ifthenelse{\equal{#2}{}}{\ProbOpr{VAR}_{#1}}
{\ifthenelse{\equal{#1}{}}{\ProbOpr{VAR}\left[#2\right]}{\ProbOpr{VAR}_{#1}\left[#2\right]}}} %

\newcommand{\vo}{{\vct{o}}}

\newcommand{\eat}[1]{}

\definecolor{egogreen}{RGB}{214,235,197}
\definecolor{refred}{RGB}{251,210,204}
\definecolor{LightCyan}{rgb}{0.88,1,1}
\definecolor{LightRed}{rgb}{1,0.94,0.94}
\definecolor{LightBlue}{rgb}{0.94,0.97,1}
\definecolor{scarlet}{RGB}{147,0,0}
\definecolor{citeblue}{RGB}{238,26,28}
\definecolor{Blue}{RGB}{51,10,154}

\definecolor{citecolor}{RGB}{50, 63, 138}
\definecolor{linkcolor}{RGB}{187,18,26}

\newcommand{\ours}{AutoReg3D}

\definecolor{mymagenta}{HTML}{FF00FF}
\definecolor{myblue}{HTML}{72D0Fb}

\definecolor{cvprblue}{rgb}{0.21,0.49,0.74}
\usepackage[pagebackref,breaklinks,colorlinks,allcolors=cvprblue]{hyperref}

\def\paperID{****} 
\def\confName{CVPR}
\def\confYear{2026}

\title{On the Feasibility and Opportunity of Autoregressive 3D Object Detection}

\author{
Zanming Huang\textsuperscript{1} \;
Jinsu Yoo\textsuperscript{1} \;
Sooyoung Jeon\textsuperscript{1} \;
Zhenzhen Liu\textsuperscript{2} \\
Mark Campbell\textsuperscript{2} \;
Kilian Q Weinberger\textsuperscript{2} \;
Bharath Hariharan\textsuperscript{2} \;
Wei-Lun Chao\textsuperscript{1,3} \;
Katie Z Luo\textsuperscript{4}\;
\\
\\
\textsuperscript{1}The Ohio State University~~~\textsuperscript{2}Cornell University~~~\textsuperscript{3}Boston University~~~\textsuperscript{4}Stanford University \\
}

\begin{document}
\maketitle
\begin{abstract}

LiDAR-based 3D object detectors typically rely on proposal heads with hand-crafted components like anchor assignment and non-maximum suppression (NMS), complicating training and limiting extensibility. We present \ours{}, an autoregressive 3D detector that casts detection as sequence generation. Given point-cloud features, \ours{} emits objects in a range-causal (near-to-far) order and encodes each object as a short, discrete-token sequence consisting of its center, size, orientation, velocity, and class. This near-to-far ordering mirrors LiDAR geometry---near objects occlude far ones but not vice versa---enabling straightforward teacher forcing during training and autoregressive decoding at test time. \ours{} is compatible across diverse point-cloud or backbones and attains competitive nuScenes performance without anchors or NMS. Beyond parity, the sequential formulation unlocks language-model advances for 3D perception, including GRPO-style reinforcement learning for task-aligned objectives. These results position autoregressive decoding as a viable, flexible alternative for LiDAR-based detection and open a path to importing modern sequence-modeling tools into 3D perception.

\end{abstract}    
\section{Introduction}
\label{sec:intro}

\begin{figure}[!t]
\centering
\includegraphics[width=0.9\linewidth]{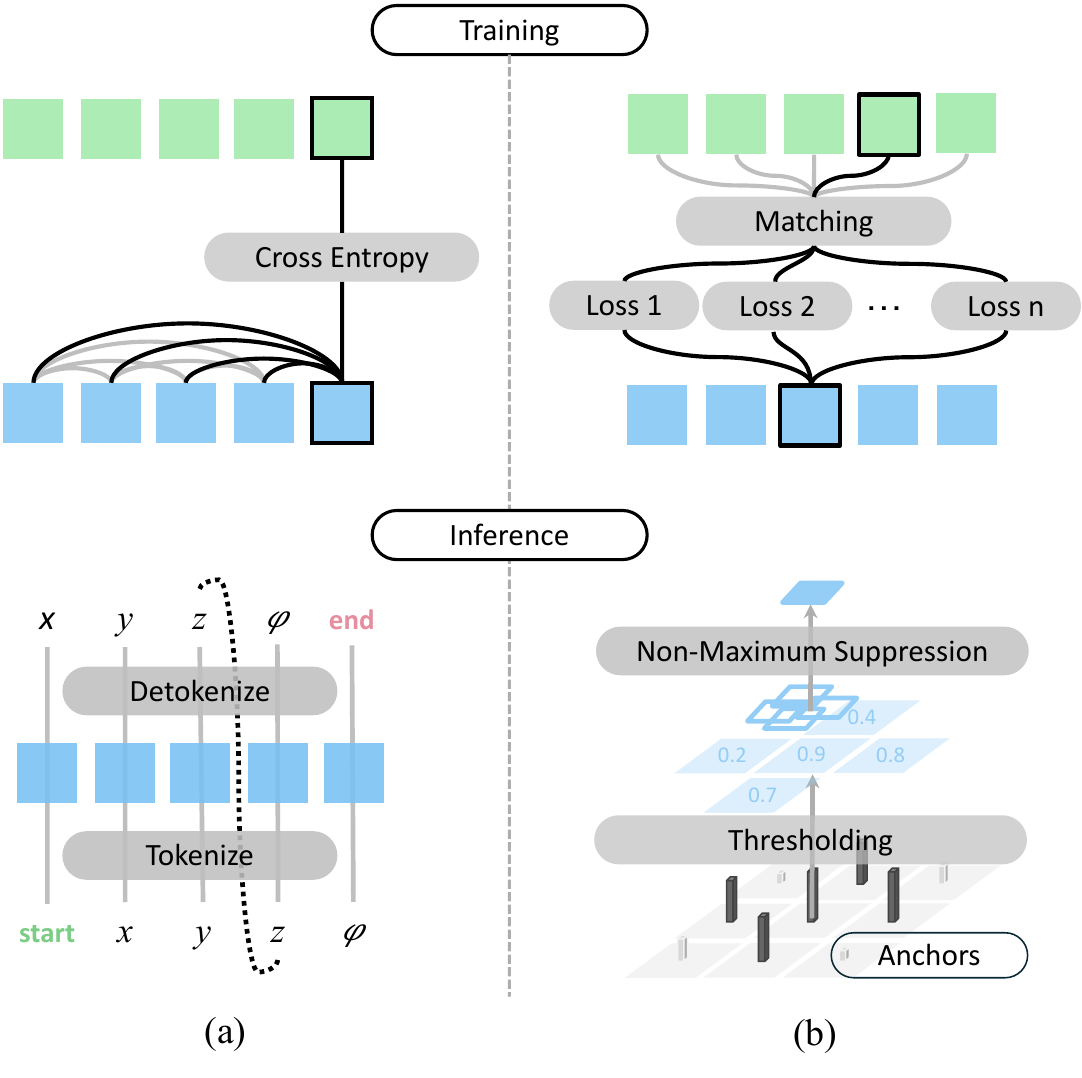}

\vspace{-0.5em}
\caption{\textbf{Autoregressive Object Detection for 3D.} Our work proposes a 3D object detector that leverages a sequential generation representation (a). This eliminates many of the complications associated with the rigid detection pipeline, including anchor assignment, confidence thresholding, and NMS (b).}
\label{fig:teaser}

\vspace{-0.5em}
\end{figure}

Object detection has long been one of the most influential tasks in computer vision, with applications spanning recognition \cite{ren2015faster,redmon2016you,redmon2017yolo9000}, healthcare \cite{liu2019privacy,ragab2024comprehensive}, and a wide range of analytics \cite{buric2018object,van2018you,jain2019evaluation}. In 3D perception—critical for scene understanding and downstream tasks such as autonomous driving and robotic systems—detectors typically follow a ``propose-then-classify'' paradigm. These systems first generate region proposals and then refine and classify them, a design directly inspired by 2D methods~\cite{girshick2014rich}. Both two-stage variants~\cite{girshick2014rich,ren2015faster,shi2019pointrcnn,shi2020pv,shi2023pv} and their single-stage counterparts~\cite{redmon2016you,redmon2017yolo9000,feng2021tood} rely on classification and regression objectives to produce the final bounding boxes, achieving strong performance and remaining the dominant approach.
 
However, proposal-based detectors depend on a rigid stack of hand-crafted machinery: anchor assignment, proposal matching, geometric regression targets, confidence thresholds, and NMS~\cite{girshick2014rich,hosang2017learning}. This stack exists because predictions are made independently across spatial locations, yielding multiple overlapping boxes that must be filtered and de-duplicated. Although effective for redundancy control, these components introduce extra complexity, complicate training, and often discard information during post-processing. They also hinder composability with downstream modules, \eg, large language models (LLMs), limiting the scalability and extensibility of 3D detection beyond the regimes for which these systems were designed.

We propose to address these issues by reintroducing dependence among predictions. Inspired by the success of autoregressive~(AR) sequence modeling in language \cite{brown2020language,team2023gemini} and emerging vision-generation tasks \cite{alayrac2022flamingo,liu2023visual}, we explore an alternative formulation for 3D detection: \emph{casting bounding-box prediction as autoregressive sequence generation}. Unlike proposal-based detectors, an autoregressive model emits one object at a time while conditioning on previously predicted objects. This dependency lets the detector remain aware of prior outputs, naturally suppressing overlaps and obviating NMS or proposal filtering. While related ideas have shown promise for 2D detection \cite{chen2021pix2seq,chen2022unified,jiang2025detect}, a scalable and competitive autoregressive alternative for 3D point-cloud detection has remained elusive due to the higher dimensionality, the challenges of discretizing continuous geometry, and the sheer spatial scale of LiDAR scenes.

We present \ours{}, the first autoregressive 3D point-cloud detector that achieves performance competitive with state-of-the-art proposal-based and query-based systems. Our key insight is that LiDAR naturally induces a near-to-far ordering: objects closer to the ego-vehicle are physically encountered---and thus observed---before those farther away. This ordering reflects the causal structure of occlusion in 3D and provides an intuitive \emph{sequential dependence} along which objects can be generated autoregressively. In contrast to 2D images, where the decoding order is largely arbitrary, 3D range provides a principled sequence axis that aligns with autoregressive modeling.

Concretely, given a point cloud encoded by any encoder (\eg,voxel-based~\cite{yan2018second}), \ours{} autoregressively generates a sequence of discrete tokens, each representing a single object. Every object is encoded as a short token sequence for its class, location $(x, y, z)$, size $(l, w, h)$, velocity $(v_x, v_y)$, and orientation $\psi$. To discretize continuous geometry effectively while respecting the distinct ranges and semantics of each axis in 3D,
we use ego-relative coordinates with parameter-specific vocabularies, expanded along each of the bound box semantic axis, \ie directions, orientation, \etc. Generation begins with a \texttt{[start]} token, proceeds from near to far, and terminates with a \texttt{[end]} token---producing a threshold-free set of boxes and eliminating confidence thresholds and NMS.

Our formulation not only simplifies 3D detection but also unlocks new capabilities. Because inference is autoregressive, \ours{} can readily adopt techniques from sequence and language modeling, including reinforcement-learning-based fine-tuning~\cite{ouyang2022training,pinto2023tuning} and advanced decoding~\cite{freitag2017beam,holtzman2019curious}.
These techniques offer straightforward, plug-in paths to further improve performance without redesigning the core model.

The goal of this work is to show viability: autoregressive modeling for 3D point-cloud detection can \emph{match mainstream accuracy} while {opening a path to modern sequence-modeling advances} in 3D. We acknowledge a practical bottleneck---sequential decoding latency---which applies broadly to AR applications, and for this work, we view speed as orthogonal to our core contribution. We expect improvements with advances in autoregressive decoding and hardware acceleration; because \ours{} uses the canonical AR toolkit, these gains should transfer with minimal integration cost. 
Altogether, our contributions are: 
\begin{itemize}
    \item We introduce \ours{}, the first autoregressive 3D object detector that directly generates object sequences from point clouds, achieving performance on par with leading proposal-based and query-based detectors.
    \item We present a detailed ablation study of design factors—including object-level tokenization, sequence ordering, and decoding methodologies—that are critical for effective autoregressive 3D detection.
    \item We demonstrate unique capabilities enabled by the autoregressive formulation, including the elimination of NMS, compatibility with reinforcement learning (RL) fine-tuning, and promptable decoding.
\end{itemize}

\section{Related Works}
\label{sec:related}

\paragraph{3D Object Detection from Point Clouds.}
3D object detection has been a cornerstone objective in the self-driving field, where methods have been developed to perceive from camera inputs \cite{girshick2015fast,ren2015faster, redmon2016you, redmon2017yolo9000,carion2020end} and 3D LiDAR point clouds \cite{lang2019pointpillars,yan2018second,shi2019pointrcnn,yin2021center,liu2022bevfusion}.
Of these, LiDAR based 3D object detectors have to contend with high number of sensor reading points, and many works have been dedicated to processing them into deep learning features regressing the bounding boxes.
These have collected into three main paradigm, consisting of point-based methods \cite{qi2017pointnet, qi2017pointnet++,yang2018pixor,yang2019std, shi2019pointrcnn,chen2022mppnet}, pillar-based methods \cite{yan2018second,lang2019pointpillars,shi2020pv,yin2021center}, and voxel-based methods \cite{shi2020pv, deng2021voxel,chen2023voxelnext,shi2023pv}. Point-based backbones were developed to directly process point cloud data, but suffer from high input scalability and have slower inference times \cite{yan2018second,yang2018pixor}. In contrast, voxel and pillar-based backbones divide the space into a constant number of features, and trade off some accuracy resolution, but are often have faster inference speeds. More recently, enabled by improved hardware, there have been improved Transformer and Mamba point cloud backbones \cite{wang2023dsvt,zhang2024voxel,liu2024lion}. 
These methods all leverage the propose-then-classify subtasks, inspired from the object detection task from the 2D counterparts \cite{girshick2014rich,girshick2015fast,redmon2016you}. Analogous to the image-based object detectors, 3D object detector heads often employ a two-stage \cite{yang2018pixor,shi2019pointrcnn,shi2020pv} or one-stage \cite{yin2021center,wang2023dsvt,liu2024lion} detection head, to name a few. 
These methods have been well refined, but due to the inherent independent nature of the propose-then-classify objective, suffer from multiple independent predictions per-location, and must be post-processed by threshold selection and suppression techniques \cite{hosang2017learning,liang2018design,yang2019learning}. Most popular among them is NMS \cite{viola2004robust,girshick2014rich}, but these methods all lead to information loss.
 
\paragraph{Autoregressive Object Detection.} Sequential generation originally was popularized within Natural Language Processing (NLP)~\citep{bengio2003neural,mikolov2010recurrent,radford2019language}, where transformer-based~\citep{vaswani2017attention} autoregressive models demonstrate great success in diverse NLP applications, such as machine translation~\citep{vaswani2017attention}, question answering~\citep{raffel2020exploring,brown2020language}, and reasoning~\citep{anil2023palm,bai2023qwen,dubey2024llama,team2025gemma}.
Recently, there has been significant interest in extending this sequential modeling paradigm to the vision domain. One direction is developing large vision-language models (VLMs)~\citep{li2023blip,liu2023visual,hurst2024gpt}. These models perform sequential language generation conditioned on visual inputs, primarily addressing tasks like visual question answering~\citep{antol2015vqa} and image captioning~\citep{vinyals2015show}. For the object detection task, Pix2Seq \cite{chen2021pix2seq} pioneered in formulating 2D object detection as a sequence prediction problem. It represents the set of bounding boxes and class labels as a discrete sequence of tokens, which a decoder generates autoregressively conditioned on the input image. Subsequent works like Pix2Seq v2~\citep{chen2022unified} and Rex-Omni ~\citep{jiang2025detect} further extend this paradigm to create unified models that can perform both object detection and a variety of other 2D vision tasks. 
For the 3D LiDAR point cloud modality, autoregressive approaches to object detection remain largely unexplored. One related work, Point2Seq~\citep{xue2022point2seq}, models detection as the sequential prediction of bounding box attributes. However, its autoregressive process only operates along the attribute dimension. Across boxes, all bird's-eye-view (BEV) cells predict for the same type of attribute in parallel. To the best of our knowledge, we are the first to investigate fully autoregressive detection for LiDAR point cloud data.

\section{Method}
\label{sec:method}

\begin{figure*}[t]
    \centering
    \includegraphics[width=\linewidth]{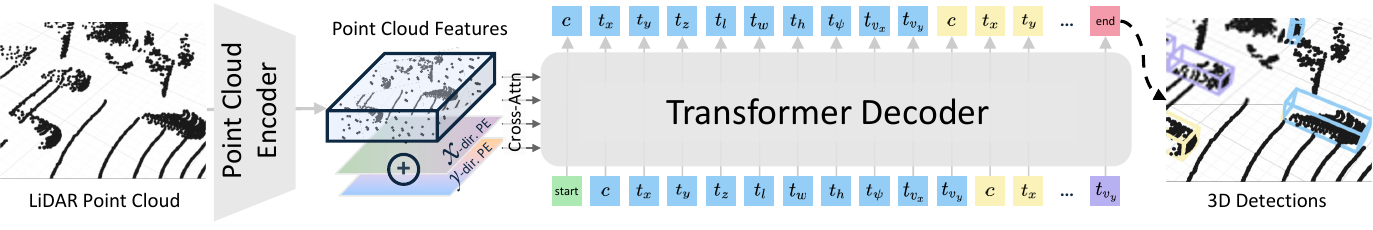}
    \vspace{-0.5em}    
    \caption{\textbf{Model Architecture.} We leverage an encoder-decoder architecture for encoding point cloud features, then generate tokenized bounding boxes with a causal Transformer decoder. We detokenize the generated sequence to obtain the final set of 3D object detections. This design is compatible with a variety of point cloud encoders, including pillar-, voxel-convolutions, transformer, and Mamba backbone.}
    \label{fig:model}
    
    \vspace{-0.5em}
\end{figure*}

Existing regression-based 3D detectors rely heavily on hand-crafted components such as customized losses for each box attribute and anchor assignment strategies~\cite{yan2018second,shi2019pointrcnn,shi2020pv,shi2023pv}. Inspired by language modeling, we cast 3D object detection as a sequence generation task, demonstrating that such complexity can be largely avoided. Our framework adopts a unified cross-entropy loss across all tokenized box attributes, and removes the need for post-processing steps such as matching~\cite{kuhn1955hungarian} and NMS, thereby significantly simplifying the modeling pipeline. Despite its simplicity, \ours{} achieves performance on par with state-of-the-art regression-based 3D detectors. Moreover, \ours{} is compatible with existing 3D backbones, enabling straightforward integration with current architectures. 

Beyond a simpler detection process, our formulation also presents unique opportunities in improving 3D detection performance. By modeling a conditional probability distribution rather than performing direct regression, we are able to unlock two key capabilities. 
First, we demonstrate that the model can be further improved through reinforcement learning. Additionally, we show that by providing hints at test time, our method can recover from failure cases that are challenging for conventional detectors. 

In this section, we begin by introducing our formulation, architecture, training objective, and inference process in Section~\ref{subsec:framework}, and then detail the unique opportunities in Section~\ref{subsec:opportunities}.

\subsection{3D Object Detection as Sequence Prediction}

\label{subsec:framework}

\paragraph{3D Object as Tokens.}
In LiDAR-based 3D object detection, objects are described by their class label and 3D bounding boxes parameterized by their center $(x, y, z)$, dimensions $(l, w, h)$, yaw angle $(\psi)$, and velocity $(v_x, v_y)$ in an ego-relative coordinate system. To represent bounding boxes using tokens, we are inspired by the quantization strategy of~\cite{chen2021pix2seq} and quantize the bounding box values uniformly into an integer between $t_k \in [1, n_k]$, where $k\in \{x, y, z, l, w, h, \psi, v_x, v_y\}$. Unlike~\cite{chen2021pix2seq}, which uses a shared vocabulary for all box parameters, we adopt a separate vocabulary for each parameter type to better model their distinct range and semantic meaning. Together with class label, each object in the scene is represented as a sequence of $10$ tokens $\{c, t_x, t_y, t_z, t_l, t_w, t_h, t_\psi, t_{v_x}, t_{v_y}\}$. 

\paragraph{Sequence Ordering.}
We order the token sequences of individual objects into a single sequence to represent the entire scene. In 2D object detection, autoregressive methods order objects randomly, and enforcing spatial ordering does not benefit final detection performance~\cite{chen2021pix2seq}. 
In contrast, objects are inherently ordered in 3D space. For instance, in ego-relative settings such as driving, nearby objects occlude those farther away. It is therefore natural to detect foreground objects first before reasoning about occluded ones. This intrinsic dependency in 3D object detection motivates the use of a deterministic ordering between objects in the scene. In particular, we arrange objects in \textit{near-to-far} order according to their distance from the ego vehicle.

\paragraph{Autoregressive Model.}
We model the joint probability distribution of all tokens in the scene given the point-cloud $\mathbf{X}$:
\begin{align}
p(\vo|\mathbf{X})=\prod_{i=1}^{N}p(\vo_i|\vo_{1:i-1}, \mathbf{X})
\end{align}
where $\vo$ denotes the sequence of object tokens. To model the conditional probability $p(\vo_i|\vo_{1:i-1}, \mathbf{X})$, we adopt an encoder-decoder architecture, the encoder extracts a global feature representation from the point cloud, while a Transformer decoder~\cite{vaswani2017attention} autoregressively predicts tokens one at a time. By conditioning each prediction on prior outputs, the model naturally captures dependencies among objects in the scene. Our approach is flexible, as it is compatible with any point cloud encoder that outputs hidden scene representations. We intentionally keep our architecture design minimal and modular, enabling integration with a wide range of existing 3D backbones.

\paragraph{Training Objective.}
We train the model to maximize the likelihood of the ground-truth token sequence $\vo$ given the input point cloud $\mathbf{X}$. The optimization objective can be written as follows:
\begin{align}
\mathcal{L}=\sum_{i=1}^{N}\log{P({\vo}_i|\vo_{1:i-1}, \mathbf{X})}
\end{align}
Unlike regression-based detectors that require multiple task-specific losses (e.g., for box center, size, orientation, and velocity), our approach uses a single unified cross-entropy loss across all token types. This formulation eliminates the need for hand-crafted losses and weighting, reinforcing the overall simplicity of our design. See Algorithm~\ref{alg:autoreg3d-train} for implementation of training loop in pseudocode.









\begin{algorithm}[H]
\caption{Training (Teacher Forcing)}
\label{alg:autoreg3d-train}

\begin{lstlisting}[style=py, language=python]
# Enc: point cloud encoder
# Dec: autoregressive Transformer decoder
# cross_entropy(): unified CE loss over all tokens

for X, boxes in dataloader: # point cloud and GT boxes
    F = Enc(X)
    boxes = sort(boxes) # enforce ordering
    
    seq = tokenizer.encode(boxes)
    seq_in  = seq[:-1]    
    seq_tgt = seq[1:] 

    logits = Dec(F, seq_in)             
    loss = cross_entropy(logits, seq_tgt)   
    loss.backward()
\end{lstlisting}
\end{algorithm}

\vspace{-1em}

\paragraph{Inference and Decoding.}
At inference time, object tokens are sampled sequentially according to the learned conditional distribution $p(\vo_i|\vo_{1:i-1}, \mathbf{X})$. Unlike regression-based detectors that directly output fixed numbers of bounding boxes with associated confidence scores, our model samples tokens from a learned distribution, reflecting the probabilistic nature of scene generation. This formulation allows the number of predicted objects to vary naturally and eliminates the need for anchors, confidence thresholds, or post-processing steps such as NMS.
We explore several decoding strategies commonly used in autoregressive models, including nucleus sampling~\cite{holtzman2020curious}, and deterministic approaches such as beam search~\cite{graves2012sequence}, and greedy decoding, where we simply choose the most probable token. In practice, we adopt greedy decoding for all experiments unless otherwise specified, as it is the simplest approach with minimal computational overhead and achieves performance comparable to more complex decoding methods. The pseudocode for inference procedure is shown in Algorithm~\ref{alg:autoreg3d-infer}.





\begin{algorithm}[H]
\caption{Inference (Autoregressive Decoding)}
\label{alg:autoreg3d-infer}

\definecolor{codeblue}{rgb}{0.25,0.5,0.5}
\lstset{
  backgroundcolor=\color{white},
  basicstyle=\fontsize{7.2pt}{7.2pt}\ttfamily\selectfont,
  columns=fullflexible,
  breaklines=true,
  captionpos=b,
  commentstyle=\fontsize{7.2pt}{7.2pt}\color{codeblue},
  keywordstyle=\fontsize{7.2pt}{7.2pt},
}

\begin{lstlisting}[style=py, language=python]
# Enc: point cloud encoder
# Dec: autoregressive Transformer decoder

def detect(X):
    F = Enc(X) 
    seq = ["<START>"] 
    while True:
        logits = Dec(F, seq) 
        t = argmax(logits[-1])  # greedy decoding
        seq.append(t)
        if t == "<END>" or len(seq) > T_max:
            break
    boxes = tokenizer.decode(seq) # tokens to 3D boxes
    return boxes
\end{lstlisting}
\end{algorithm}
\vspace{-1em}

\paragraph{Modeling Details.}
We model our detector as an encoder-decoder architecture \cite{raffel2020exploring}, where the autoregressive 3D object detector's decoder is a 6-layer Transformer decoder that cross-attends to the point-cloud features encoded by the backbone model \cite{lang2019pointpillars, yin2021center, yan2018second, liu2024lion}. A schematic of the encoder-decoder model is in \autoref{fig:model}. The cross-attention mechanism ensures that object generation remains conditioned on the input point-cloud features throughout the decoding process, following standard practice in encoder-decoder architectures. We use learnable positional embeddings for the relative point-cloud features in bird's-eye view. Specifically, we decompose the 2D position into separate embeddings for the relative indices along the $x$ and $y$ directions, then sum these two embeddings to obtain the final positional encoding for each feature. This approach provides spatial awareness by distinguishing between the two spatial dimensions while maintaining linear space complexity rather than quadratic. We further embed the position of the sequence generation with a learnable positional embedding, common in sequence modeling \cite{vaswani2017attention}. We purposefully keep the architecture design simple, and due to computational limitations, leave scaling model size to future work.

\subsection{Opportunities in Autoregressive 3D Detection}
\label{subsec:opportunities}
\paragraph{Reinforcement Learning Fine-tuning.}
While teacher-forcing maximizes token likelihood, it does not explicitly optimize the set-level detection objective. 
Our sequence prediction formulation enables RL fine-tuning of \ours{} with a sequence-level reward aligned with detection quality at inference time, thereby improving global consistency.
Specifically, we adopt RL strategy based on GRPO~\cite{shao2024deepseekmath}. Given a scene point cloud $\mathbf{X}$, we sample a group of $G$ detection sequences $\{\vo_1, \vo_2, \dots, \vo_G\}$ from the current detection policy $\pi_{\theta}$. 

We define a task-aligned reward. Given a set of ground-truth boxes $B_c=\{b_1, b_2, \dots, b_3 \}$ and a set predicted boxes $\hat{B}_c=\{\hat{b}_1, \hat{b}_2, \dots, \hat{b}_3 \}$ that belong to class $c \in \mathcal{C}$, 
we calculate each ground-truth box's maximum IoU $r^{*}_i$ with the predicted boxes in the same class.
We calculate the class-averaged reward $r$ inspired by F1:
\begin{align}
r_c^{\mathrm{Rec.}} &= \frac{1}{|B_c|} \sum_{i=1}^{|B_c|} r_i^{*}, \quad
r_c^{\mathrm{Prec.}} = \frac{1}{|\hat{B}_c|} \sum_{i=1}^{|B_c|} r_i^{*}, \nonumber \\[4pt]
r &= \frac{1}{|\mathcal{C}|} \sum_{c \in \mathcal{C}}
     \frac{2 \, r_c^{\mathrm{Prec.}} \, r_c^{\mathrm{Rec.}}}
          {r_c^{\mathrm{Prec.}} + r_c^{\mathrm{Rec.}}}.
\label{eq:reward}
\end{align}
The GRPO objective is formulated as follows:
\vspace{-1em}
\begin{align}
\mathcal{J}_{\text{GRPO}}(\theta)
&= \frac{1}{G} \sum_{i=1}^{G} \frac{1}{|\vo_i|} 
   \sum_{t=1}^{|\vo_i|} \Big[ \nonumber \\
&\quad \min\!\Big(
      \rho_{i,t} \hat{A}_{i,t},\,
      \text{clip}(\rho_{i,t}, 1 - \epsilon, 1 + \epsilon)\hat{A}_{i,t}
    \Big) \nonumber \\
&\quad -\, \beta\, \mathbb{D}_{\mathrm{KL}}\!\left[
      \pi_\theta \,\big\|\, \pi_{\text{ref}}
    \right] \Big]
\label{eq:grpo}
\end{align}
Where $\hat{A}_{i,t}$ is the estimated advantage, $\rho_{i,t}$ is the importance sampling ratio, and $\pi_\text{ref}$ is the frozen detection policy after supervised training.

\paragraph{Cascading Refinement.}
A notable advantage of our conditional autoregressive formulation is its ability to incorporate external inputs as hints during inference. Unlike regression-based detectors, which predict objects independently, our model can condition future predictions on user-provided or pre-existing information, such as bounding boxes or partial detections, as input tokens to guide subsequent predictions.

To illustrate this capability, we cascade the output of a prior model, which generates a strong initial prediction, into a completion model that refines it. We then aggregate the predictions using a simple clustering strategy, where overlapping boxes with IoU above a threshold are aggregated. This procedure yields consistent improvements over using either the prior model or the completion model alone.

Overall, this conditional reasoning capability highlights the potential of the sequence-based formulation beyond standard 3D detection, allowing the model to incorporate external priors or partial observations from other detectors, systems, or user inputs.

\section{Experiments}
\label{sec:exp}

In this section, we compare \ours{} with state-of-the-art LiDAR-based 3D detectors on nuScenes~\cite{nuscenes}, demonstrate unique capabilities enabled by autoregressive formulation (RL-based optimization and interactive correction), and present ablation studies analyzing key design choices.
\subsection{Experiment Setup}
We first describe the experimental setup, including the dataset, evaluation metrics, and implementation details.

\paragraph{Dataset.} 
We conduct our experiments on the nuScenes dataset~\cite{nuscenes}. nuScenes is a large-scale autonomous driving dataset that is widely used for 3D perception. It consists of $1{,}000$ driving scenes collected in Boston and Singapore, with $700$ scenes for training,
$150$ for validation, and $150$ for testing. The dataset provides 3D annotations for $10$ object categories,
recorded at $2\,\mathrm{Hz}$ and covering a LiDAR range of approximately $50\,\mathrm{m}$.

\paragraph{Metrics for Detection.}
The standard nuScenes 3D detection benchmark reports mean Average Precision (mAP) and the nuScenes Detection Score (NDS). NDS is computed as a weighted combination of mAP and five true-positive error terms that measure translation, scale, orientation, velocity, and attribute accuracy. 
However, mAP requires detectors to output confidence scores, which is natural for traditional anchor-based 3D detectors. These scores originate from formulating detection as a classification problem, where each anchor predicts how likely it corresponds to an object versus background. In contrast, our method casts detection as a sequence generation problem and does not produce such objectness scores. As a result, the standard mAP and NDS metrics are not directly applicable.
Similar to \cite{jiang2026rexomni}, we adopt Precision, Recall, and F1 score as our evaluation metrics, which are score-independent. Following nuScenes conventions, a prediction is considered correct if its center lies within a set of official distance thresholds from a ground-truth box. For each class and each distance threshold, we compute Precision, Recall, and F1, and report their average.

\paragraph{Tokenization.}
To build our customized vocabulary, we uniformly quantize each continuous box parameter $x, y, z, l, w, h, \psi, v_x, v_y$. Specifically, we determine bin count
by balancing quantization error with vocab size, with bin widths $0.05\,\mathrm{m}$ for center/size; $0.05\,\mathrm{rad}$ for yaw; and $0.1\,\mathrm{m/s}$ for velocity.
Together with class tokens and the special \texttt{[start]}, \texttt{[end]}, and \texttt{[pad]} tokens, this results in a vocabulary of $6{,}819$ tokens. 
More details and analysis of different tokenization strategies are provided in \cref{subsec:supp_tokenization} of the supplementary.

\paragraph{Implementation Details.} 
For supervised learning (\ie, teacher forcing), we adopt the AdamW optimizer with a learning rate of $1\times 10^{-3}$ and a cosine warm-up and decay schedule. We follow the standard data processing and augmentation pipeline commonly used in prior LiDAR-based 3D detectors~\cite{bai2022transfusion,liu2024lion,wang2023dsvt}. During autoregressive decoding, we restrict sampling to the subset of tokens valid for the current attribute type.
For GRPO training, we use a group size of 8, omit the KL penalty by setting $\beta=0$, and train with a batch size of 64. Please refer to supplementary for more implementation details.

\subsection{Main Results}

\begin{table}[t]
    \small
    \centering
    \begin{adjustbox}{width=\linewidth, center}
        \begin{tabular}{l c c c c c}
            \toprule
            Method & Encoder & Det. Head & Prec. & Rec. & F1 \\
            \midrule
            PointPillars~\cite{lang2019pointpillars} 
                & \multirow{3}{*}{Pillar Conv.} 
                & Anchor-based 
                & 58.3 & 50.0 & 53.1 \\

            CenterPoint~\cite{yin2021center} 
                &  
                & Center-based 
                & 67.9 & 53.3 & 59.5 \\

            Ours 
                &  
                & AR Transformer 
                & 69.6 & 52.4 & 59.2 \\

            \midrule

            SECOND~\cite{yan2018second} 
                & \multirow{3}{*}{Voxel Conv.} 
                & Anchor-based 
                & 63.5 & 55.6 & 59.1 \\

            CenterPoint~\cite{yin2021center} 
                &  
                & Center-based 
                & 72.8 & 60.3 & 65.8 \\

            Ours 
                &  
                & AR Transformer 
                & 74.9 & 59.4 & 65.8 \\

            \midrule

            DSVT~\cite{wang2023dsvt} 
                & \multirow{2}{*}{Transformer} 
                & Non-AR Transformer 
                & 79.1 & 66.3 & 71.6 \\

            Ours 
                &  
                &  AR Transformer
                & 77.0 & 64.1 & 69.5 \\

            \midrule

            LION~\cite{liu2024lion} 
                & \multirow{2}{*}{Mamba} 
                &  Non-AR Transformer
                & 78.6 & 68.3 & 72.5 \\

            Ours 
                &  
                &  AR Transformer
                & 77.5 & 65.2 & 70.4\\

            \bottomrule
        \end{tabular}
    \end{adjustbox}

    \vspace{-0.5em}    \caption{\textbf{NuScenes Validation Detection Performance.} We report precision, recall, and F1 results for \ours{} compared to baseline methods, grouped by encoder type. For methods that require thresholding, we select the threshold that yields the highest F1 score on the training set. Across all encoder types, \ours{} achieves competitive performance and surpasses proposal-then-classify detectors.}
    \label{tab:main_result}

    \vspace{-0.3em}
\end{table}

\begin{figure}[t]
\centering
\includegraphics[width=\linewidth]{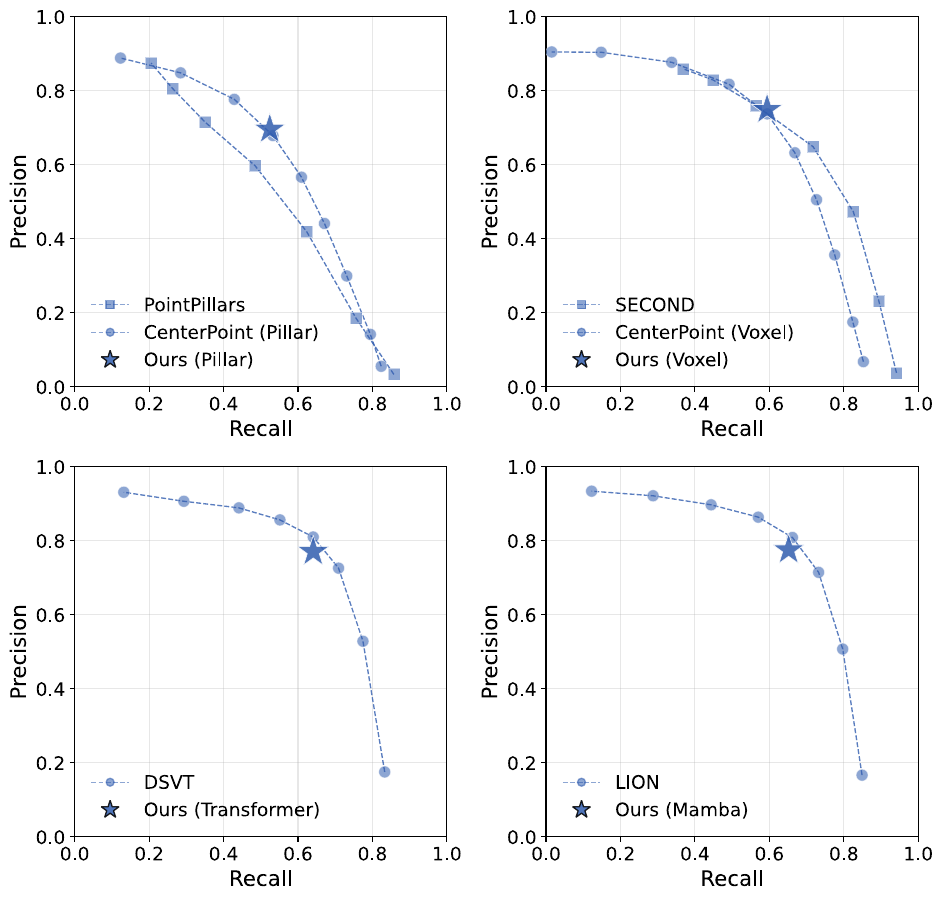}

\vspace{-1em}
\caption{\textbf{Precision-Recall Plot.} We plot the PR curves for the baseline methods, and the precision-recall point using our autoregressive decoder with a star. \textit{Top left}: Pillar-based backbone. \textit{Top right}: Voxel-based backbone. \textit{Bottom left:} Transformer-based backbone. \textit{Bottom right:} Mamba-based backbone. 
We observe that the precision-recall point of \ours{} consistently hits or lies outside the PR curves of models with the same backbone.
}
\label{fig:pr_curve}

\vspace{-1.5em}
\end{figure}

\paragraph{Supervised Training Performance.}
We evaluate the feasibility and general adaptability of our autoregressive formulation by comparing our method against a diverse set of commonly adopted baseline methods spanning three representative point cloud encoder families: pillar-based, voxel-based, transformer-based, and Mamba-based architectures. For each encoder type, we initialize our model with the corresponding pre-trained backbone used from prior works. Specifically, we use the encoder weights from CenterPoint-Pillar~\cite{yin2021center} for pillar-based models, CenterPoint-Voxel for voxel-based models, DSVT~\cite{wang2023dsvt} for transformer-based models , and LION~\cite{liu2024lion} for Mamba-based models.

\begin{figure*}[ht]
    \centering
    \includegraphics[width=\textwidth]{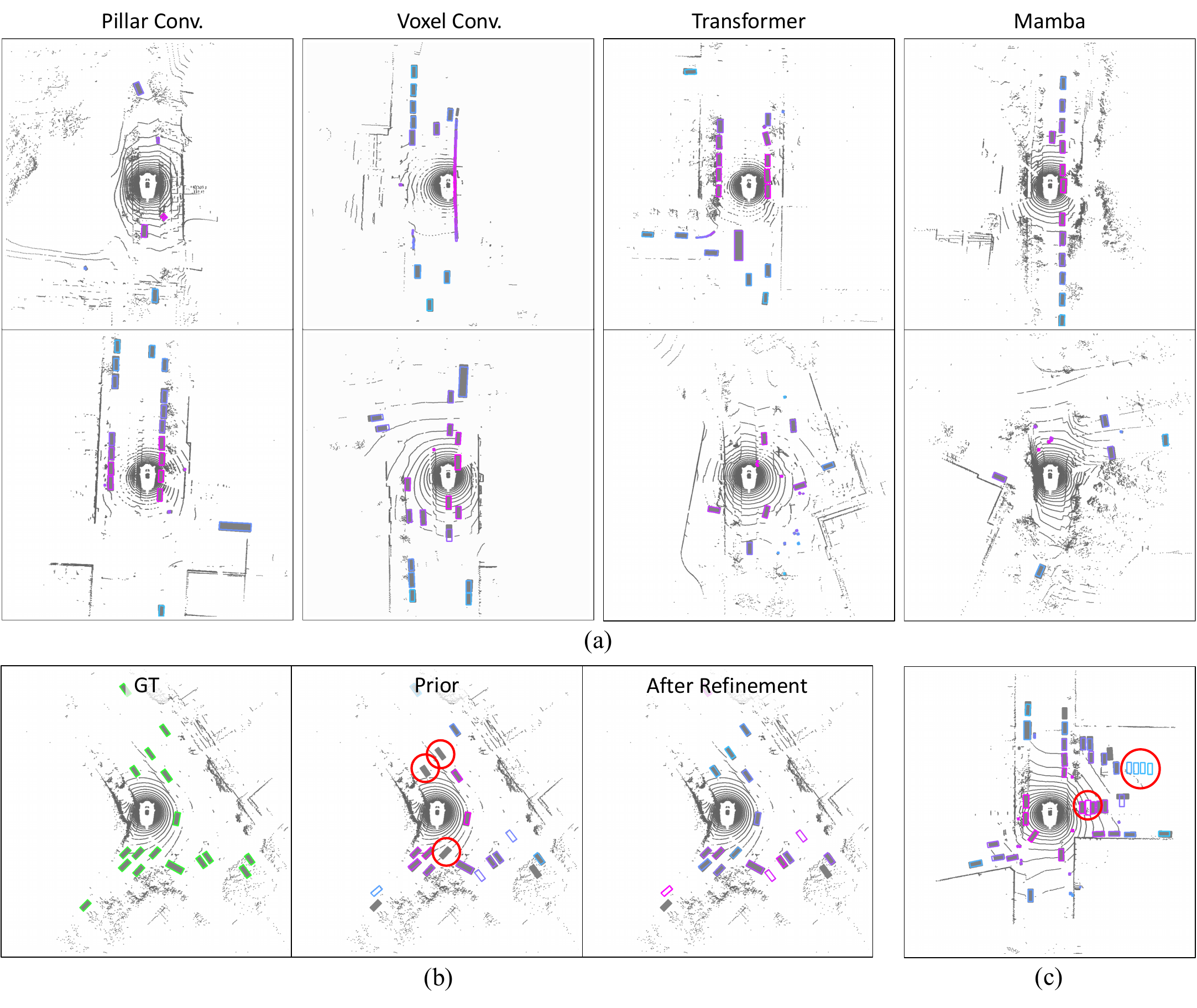}
    \vspace{-2.5em}    
    \caption{\textbf{Qualitative Results.} (a) Bounding box generations from our method across four different encoder backbones; (b) Cascading refinement visualization with ground-truth boxes (left, outlined in green), predictions from the near-to-far prior model (middle), and resulting predictions (right). Cascading Refinement recovers objects missed by the prior model (circled in red). (c) Failure case example. \ours{} generates boxes from first (\textcolor{mymagenta}{magenta}) to last (\textcolor{myblue}{blue}), ground-truth boxes are in gray. Best viewed in color.}
    \label{fig:qualitative_example}
    \vspace{-0.5em}
\end{figure*}

Across all encoder types, \ours{} achieves precision–recall near the precision–recall frontier of their respective baselines (\autoref{fig:pr_curve}). For fair comparison, we select the confidence thresholds for all baseline detectors by choosing the value that maximizes F1 score on the \textit{training} set. As shown in \autoref{tab:main_result}, our method achieves performance on par with existing regression-based detectors across all encoder types. Our voxel-based model matches the F1 score of CenterPoint~\cite{yin2021center} at 65.8. 
Notably, for both pillar-based and voxel-based models, our method attains higher precision than their regression-based counterparts. We attribute this to the interdependent nature of our autoregressive generation process, where each box is conditioned on previously generated ones. This reduces false positives compared to regression-based methods that assume independence among objects. See \autoref{fig:qualitative_example}\,(a) for qualitative examples.

\paragraph{Reinforcement Learning Performance.}
Our autoregressive formulation also enables further performance gains through reinforcement learning. During RL fine-tuning, we freeze the encoder and optimize only the autoregressive detection head using GRPO. As shown in \autoref{tab:rl}, this additional training stage improves the voxel-based model’s F1 score from 65.8 to 66.7. The improvement is primarily driven by increased recall, reflecting the impact of our task-specific reward, which encourages the model to generate more complete prediction sequences and successfully detect objects that were previously missed.

\begin{table}[t]
    \centering
    \small
    \begin{adjustbox}{width=0.85\linewidth, center}
        \begin{tabular}{lccccc}
            \toprule
            Model~~~~~~~~~~~~~~~~~~~~~~~~~~~~~~ & Precision & Recall & F1 \\
            \midrule
            Teacher Forcing                & \textbf{74.9} & 59.4 & 65.8 \\
            + GRPO      & 74.5 & \textbf{60.9} & \textbf{66.7 }\\
            \bottomrule
        \end{tabular}
    \end{adjustbox}
    
    \vspace{-0.5em}
    \caption{\textbf{Performance with RL Fine-tuning.} Observe that fine-tuning with GRPO directly on IoU further boosts performance.}
    \label{tab:rl}

    \vspace{-0.5em}
\end{table}

\subsection{Method Analysis}

We perform ablation on token ordering, decoding and inference method using our voxel-based autoregressive model.

\paragraph{Effect of Object Ordering.}
We study how different object ordering strategies affect detection performance in our autoregressive framework. As shown in \autoref{tab:obj_order}, we compare three orderings: random order, descending order by the number of LiDAR points within each ground-truth box, and a distance-based near-to-far ordering. We observe that the near-to-far ordering significantly outperforms both random and point number-based ordering. By predicting objects from near to far, the model effectively exploits the interdependent nature of the task, where close-by objects inform the prediction of farther objects. Ordering by point count performs better than random ordering, as it partially correlates with distance (closer objects often contain more points), but it remains an imperfect proxy because small yet close objects (e.g., pedestrians) may have fewer points. Random ordering performs the worst, as it fails to exploit any structural dependencies between objects.

\begin{table*}[t]
\centering
\small

\begin{minipage}[t]{0.32\textwidth}
\centering
\begin{adjustbox}{width=\linewidth}
\begin{tabular}{lccc}
\toprule
Order & Precision & Recall & F1 \\
\midrule
Random        & 68.9 & 49.9 & 56.3 \\
Point Number  & 72.8 & 55.2 & 61.8 \\
Near-to-far   & \textbf{74.9} & \textbf{59.4} & \textbf{65.8} \\
\bottomrule
\end{tabular}
\end{adjustbox}
\vspace{-0.4em}
\captionof{table}{\textbf{Ablation on Object Ordering.}}
\label{tab:obj_order}
\end{minipage}\hfill
\begin{minipage}[t]{0.305\textwidth}
\centering
\begin{adjustbox}{width=\linewidth}
\begin{tabular}{lccc}
\toprule
Order & Precision & Recall & F1 \\
\midrule
Cls. Last   & 74.4 & 58.1 & 64.9 \\
Cls. Middle & 74.8 & 58.4 & 65.2 \\
Cls. First  & \textbf{74.9} & \textbf{59.4} & \textbf{65.8} \\
\bottomrule
\end{tabular}
\end{adjustbox}
\vspace{-0.4em}
\captionof{table}{\textbf{Ablation on Token Ordering.}}
\label{tab:tok_order}
\end{minipage}\hfill
\begin{minipage}[t]{0.315\textwidth}
\centering
\begin{adjustbox}{width=\linewidth}
\begin{tabular}{lccc}
\toprule
Method & Precision & Recall & F1 \\
\midrule
Nucleus     & 67.1 & 57.8 & 61.9 \\
Greedy      & 74.9 & 59.4 & 65.8 \\
Beam Search & \textbf{75.0} & \textbf{59.9} & \textbf{66.1} \\
\bottomrule
\end{tabular}
\end{adjustbox}
\vspace{-0.4em}
\captionof{table}{\textbf{Ablation on Decoding Method.}}
\label{tab:decode}
\end{minipage}\hfill

\end{table*}

\paragraph{Effect of Token Ordering.}
We also ablate the ordering of tokens within each object sequence. As shown in \autoref{tab:tok_order}, we compare three strategies: placing the class token at the beginning, in the middle (after box location), or at the end of the sequence. Both the class-first and class-middle orderings outperform the class-last variant, with class-first yielding the best results. This indicates that having the model determine the object class earlier in the sequence provides useful context for predicting the remaining attributes, thereby improving overall detection performance.

\paragraph{Effect of Decoding Method.}
We evaluate how different decoding strategies influence detection performance, as shown in \autoref{tab:decode}. Specifically, we compare nucleus sampling~\cite{holtzman2020curious} (top-$p$=0.95, top-$k$=50), greedy decoding, and beam search~\cite{graves2012sequence} (4 beams). Beam search achieves the best performance by trading off inference time for accuracy, as it explores multiple candidate sequences and approximates a more globally optimal decoding trajectory, whereas greedy decoding selects only the most likely next token. Nucleus sampling performs the worst, as it favors diversity over accuracy, which is essential for precise object detection.

\begin{table}[t]
    \centering
    \small
    \begin{adjustbox}{width=0.85\linewidth, center}
        \begin{tabular}{lccccc}
            \toprule
            Model & Precision & Recall & F1 \\
            \midrule       
            Prior only     & \textbf{74.9} & 59.4 & 65.8 \\
            Completion only     & 68.9 & 49.9 & 56.3 \\
            \midrule
            Prior $\rightarrow$ Completion     & 74.7 & \textbf{60.2} & \textbf{66.2} \\
            \bottomrule
        \end{tabular}
    \end{adjustbox}

    \vspace{-0.5em}    \caption{\textbf{Cascading Refinement Performance.} Refinement via prompting the generation of a random-order model (Completion) improves performance over the distance-ordering model (Prior).}
    \label{tab:tts}

\end{table}

\paragraph{Results on Cascading Refinement.}
Our near-to-far model accurately detects initial objects but its distance-based ordering prevents recovery of missed boxes at arbitrary locations. The random-order model, while less accurate initially, can generate boxes anywhere in the scene. We combine their complementary strengths through conditional sampling: the near-to-far model generates initial predictions, then the random-order model produces additional detections conditioned on these results to fill gaps. As shown in \autoref{tab:tts}, this cascading approach outperforms either model alone. Notably, given approximately correct context from the distance-based model, the random-order model significantly exceeds its standalone performance and improves the performance over the initial prior. A qualitative example is shown in \autoref{fig:qualitative_example}\,(b).

\paragraph{Performance with Occlusion.}
We study the performance of \ours{} under occlusion by evaluating objects across four different visibility levels. \autoref{tab:occlusion} shows that \ours{} improves over the baseline \emph{in cases of high occlusion} (visibility $\le$ 60\%), with the largest improvement at the most occluded case (0--40\% visible). This supports the intuition that autoregressive modeling leverages the natural inter-object dependencies of 3D scenes, improving detection on partially observed objects.

\paragraph{Failure Mode Analysis.}
Similar to existing detectors, \ours{}'s failure modes include missed detections, extra detections, and misclassifications, especially for distant or sparse objects. Interestingly, the extra boxes predicted by \ours{} generally respect the overall scene structure~(\autoref{fig:qualitative_example}\,(c)): they rarely overlap with existing predictions and tend to follow road geometry. In contrast to methods that model objects independently, the autoregressive formulation leverages object inter-relations by conditioning on previous predictions, allowing \ours{} to produce plausible “guesses” grounded in scene geometry.

\begin{table}[t]
\centering
\small
\begin{adjustbox}{width=0.95\linewidth}
\begin{tabular}{lcccc}
    \toprule
    \multirow{2}{*}{Method} & \multicolumn{4}{c}{Visibility Level} \\
    \cmidrule(lr){2-5}
    & 0--40\% & 40--60\% & 60--80\% & 80--100\% \\
    \midrule
    CenterPoint & 28.9 & 30.5 & 43.2 & 67.0 \\
    Ours  & 30.0 & 31.1 & 42.7 & 67.6 \\
    $\Delta$\% & +4.1\% & +1.8\% & -1.3\% & +0.8\% \\
    \bottomrule
\end{tabular}
\end{adjustbox}
\caption{\textbf{Performance with Occlusion.} We compare F1 scores between \ours{} and the baseline under different visibility levels. Autoregressive modeling improves detection particularly under low visibility (heavy occlusion).}
\label{tab:occlusion}
\end{table}

\section{Future Works and Discussions}
\label{sec:discussions}
\vspace{-0.3em}

\paragraph{Future Directions.}
Our work focuses on demonstrating the viability of autoregressive 3D point-cloud detection.
Further experiments should explore model scaling to larger datasets and architectures \cite{dubey2024llama, Qwen-VL}. 
Additionally, while we leverage a semantically expanded vocabulary, learned codebooks \cite{van2017neural} that are specific to object detection could improve token efficiency and representation flexibility. 
These directions were beyond our computational budget but represent natural extensions.

\paragraph{Inference Speed and Applicability.}
A shared limitation of the autoregressive formulation is inference latency. Despite using bfloat-16 precision and KV-caching, our implementation achieves $\sim$5~Hz under batched inference, corresponding to 1--2~Hz for single-scene inference with voxel-based backbones. While optimization techniques from language modeling exist, their effectiveness for detection remains unexplored, particularly for the LiDAR domain.

\paragraph{Opportunities Enabled by Autoregressive Detectors.}
By formulating object detection as autoregressive modeling, we connect 3D detection to broader sequence modeling advances. Test-time scaling techniques \cite{chen2024expanding,muennighoff2025s1,ehrlich2025codemonkeys} that trade inference compute for performance may apply to detection. Furthermore, autoregressive representations could facilitate integration with language \cite{raffel2020exploring,team2023gemini,dubey2024llama} and vision-language models \cite{li2023blip,liu2023visual,Qwen-VL}. This alignment could enable spatial-linguistic reasoning and allow LLMs to leverage 3D spatial data in pretraining. However, achieving such cross-modal alignment requires addressing fundamental differences in data and task objectives not explored in this work. We view our contribution as establishing autoregressive detection viability as a foundation for these future research directions.

\section{Conclusion}
\vspace{-0.2em}
This work provides the first viability study for autoregressive 3D object detectors, demonstrating that sequence modeling can achieve competitive performance with proposal-based and query-based methods on standard benchmarks. 
By leveraging the natural near-to-far ordering in LiDAR data and discretizing geometry with parameter-specific vocabularies, we show that the rigid detection pipeline—anchor assignment, NMS, confidence thresholds—can be replaced with a single autoregressive decoder. 
Our approach establishes that 3D detection can be formulated as sequence generation, connecting it to the broader ecosystem of autoregressive modeling advances. 
We hope this work encourages further exploration of sequence modeling for 3D perception tasks.

\section*{Acknowledgment}

This work was supported by the National Artificial Intelligence Research Resource (NAIRR) Pilot and the Anvil supercomputer (award NSF-OAC 2005632). We also thank the Ohio Supercomputer Center for providing computational resources.
This work was funded in part by the National Science Foundation (IIS-2107077, IIS-2107161) and the New York Presbyterian Hospital.

{
    \small
    \bibliographystyle{ieeenat_fullname}
    \bibliography{main}
}

\clearpage
\setcounter{page}{1}
\maketitlesupplementary

\renewcommand{\thesection}{\Alph{section}}
\renewcommand{\thesubsection}{\thesection.\arabic{subsection}}
\renewcommand{\thetable}{A\arabic{table}}
\renewcommand{\thefigure}{A\arabic{figure}}
\renewcommand{\theequation}{A\arabic{equation}}
\setcounter{figure}{0}
\setcounter{table}{0}
\setcounter{equation}{0}
\setcounter{section}{0}

In this supplementary material, we provide more details and experiment results in addition to the main paper:

\begin{itemize}
    \item \cref{sec:supp_3d_head}: Discussion on 3D detection heads.
    \item \cref{sec:supp_model_arc}: Discussion on model architecture.
    \item \cref{sec:supp_cascaderefine}: Details on Cascading Refinement methodology.
    \item \cref{sec:supp_experiments}: Additional experiments.
    \item \cref{sec:supp_imp_details}: Additional details on training and inference.
    \item \cref{sec:supp_quantitative}: Additional quantitative results.
    \item \cref{sec:supp_qualitative}: Additional qualitative results.
    
\end{itemize}

\section{Discussion on 3D Detection Heads}
\label{sec:supp_3d_head}
Historically, 3D object detection heads have evolved through three main paradigms to address the challenges of accurate bounding box prediction. Anchor-based detection heads \cite{zhou2018voxelnet,yan2018second,lang2019pointpillars,shi2020pv} extend 2D detection principles by utilizing predefined 3D bounding box templates distributed across feature space, achieving strong performance but suffering from hyperparameter sensitivity regarding anchor configurations and requiring careful tuning for different object categories. 
To address these limitations, center-based detection heads emerged, directly predicting object centers and associated properties. CenterPoint \cite{yin2021center} pioneered this by detecting object centers as keypoints in bird's-eye view before regressing to 3D boxes, while FCOS3D \cite{wang2021fcos3d} extended this to monocular detection. AFDetV2 \cite{hu2022afdetv2} and FCAF3D \cite{rukhovich2022fcaf3d} demonstrated superior LiDAR-based performance without anchor matching overhead. While these methods eliminate hand-crafted anchor designs and handle varying object sizes naturally, they still require post-processing like NMS.
More recently, DETR-style heads \cite{carion2020end} leverage transformers with learnable object queries. DETR3D \cite{wang2022detr3d} introduced sparse queries that predict 3D boxes through cross-attention with multi-view features, while PETR \cite{liu2022petr} added position-embedded queries and BEVFormer \cite{li2022bevformer,Yang2022BEVFormerVA} combined this with bird's-eye view representations. These query-based methods eliminate anchors and NMS, enabling end-to-end training and modeling long-range dependencies. 
However, they typically require complex query initializations and suffer from limited architectural flexibility due to reliance on fixed query-target matching mechanisms.
In contrast, \ours{} offers a flexible autoregressive formulation, removing the complex target assignment, anchors, and post-processing thresholds, resulting in a simplified training and inference pipeline.

\section{Discussion on Model Architecture}
\label{sec:supp_model_arc}
In this work, we present a simple and straightforward encoder–decoder design for autoregressive 3D detection. A natural alternative is to directly feed point-cloud features into a decoder-only Transformer via self-attention~\cite{liu2023visual}. We conducted preliminary experiments with a decoder-only formulation using a Q-Former~\cite{li2023blip} to compress point-cloud features. However, this approach did not yield meaningful performance improvements over the simpler encoder–decoder formulation and additionally exhibited reduced training stability.

Since our goal is to demonstrate a simple and clean solution to 3D autoregressive detection, we focus on the encoder–decoder design. We leave the exploration of more complex or optimized architectural variants to future work.

\begin{figure}[ht!]
    \centering
    \includegraphics[width=\linewidth]{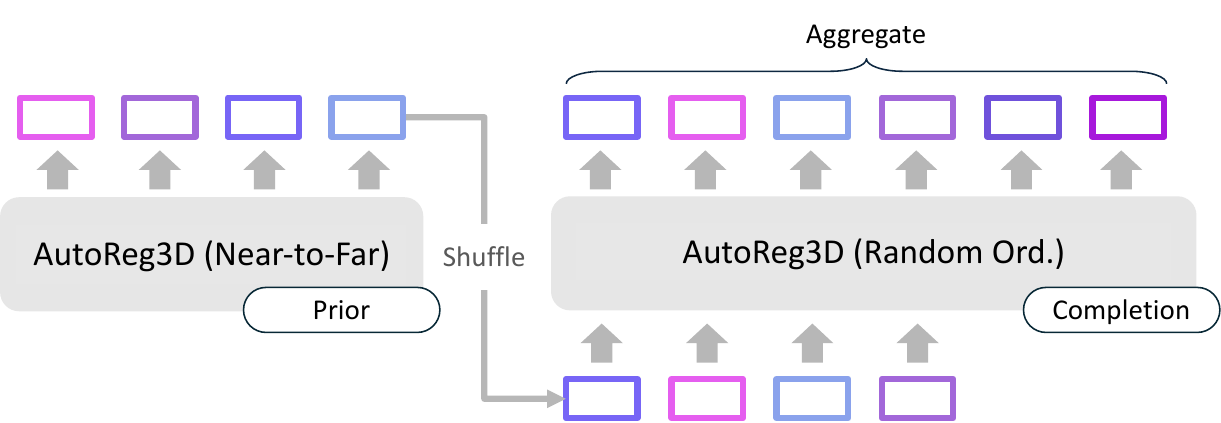}
    \caption{\textbf{Cascading Refinement Methodology.}}
    \label{fig:cascading-refinement}
\end{figure}

\section{Cascading Refinement} 
\label{sec:supp_cascaderefine}
We illustrate the methodology for \textit{\textbf{Cascading Refinement}} from Section 3.2 of the main text in \autoref{fig:cascading-refinement}. For our setup, we take the output of the distance-ordering \ours{} (``prior") model and use those as context for a random-ordering \ours{} (``completion") model after shuffling to eliminate the near-to-far positional bias. The completion model generates additional boxes conditioned on the given context. The generations are then aggregated using a simple IoU-based clustering strategy. Specifically, we group boxes that have IoU over a selected threshold, and we average the box parameters within each group. We set the IoU threshold to be 0.1. Results are reported in \autoref{tab:tts} of the main text.
Additional qualitative results for Cascading Refinement are provided in \autoref{fig:supp_cascade_1} and \autoref{fig:supp_cascade_2}.

\section{Additional Analysis}
\label{sec:supp_experiments}

\subsection{Ordering Comparison by Distance}

We analyze the performance differences between the random-order model and the near-to-far model. As shown in \autoref{tab:supp_dist_compare}, the near-to-far model outperforms the random-order model across all distance bins and all metrics. Moreover, the F1 gap between the two models grows with distance from the ego vehicle. By generating the nearby objects first, the near-to-far model is able to better reason about farther objects using previously predicted boxes as hints. In contrast, the random-order model does not leverage this structure, limiting its ability to detect distant objects, leading to a widening performance gap at larger distances. These results highlight the importance of modeling interdependencies between objects in autoregressive 3D detection.

\begin{table}[t]
    \centering
    \small
    \begin{adjustbox}{width=\linewidth, center}
        \begin{tabular}{lcccc}
            \toprule
            & 0--10m & 10--20m & 20--30m & 30m+ \\
            \midrule
            \multicolumn{5}{l}{\textit{Precision}} \\
            \midrule
            Random        & 77.5 & 72.8 & 64.1 & 49.2 \\
            Near-to-far   & 80.7 & 80.1 & 70.8 & 50.5 \\
            \midrule
            \multicolumn{5}{l}{\textit{Recall}} \\
            \midrule
            Random        & 73.5 & 62.4 & 46.5 & 25.7 \\
            Near-to-far   & 81.6 & 72.5 & 55.6 & 32.1 \\
            \midrule
            \multicolumn{5}{l}{\textit{F1}} \\
            \midrule
            Random        & 74.1 & 66.1 & 52.0 & 31.2 \\
            Near-to-far   & 79.6 & 75.8 & 61.7 & 38.2 \\
            $\Delta$ (\%) & \color{OliveGreen}{$+7.5\%$} & \color{OliveGreen}{$+14.6\%$} & \color{OliveGreen}{$+18.8\%$} & \color{OliveGreen}{$+22.7\%$} \\
            \bottomrule
        \end{tabular}
    \end{adjustbox}

    \caption{\textbf{Ordering Comparison by Distance to Ego}. Near-to-far order outperforms random order across all metrics and distance bins.}
    \label{tab:supp_dist_compare}
\end{table}

\subsection{Tokenization Methods}
\label{subsec:supp_tokenization}
We analyze different tokenization strategies for discretizing box parameters.

\paragraph{Shared \vs Expanded Vocabulary.} 
We compare a shared vocabulary across all attributes and an expanded vocabulary that assigns separate token ranges to each parameter type. In the shared-vocabulary setting, all continuous parameters are uniformly quantized into an integer between 1 and 2160. As shown in \autoref{tab:supp_vocab}, both strategies achieve similar performance, indicating that our method is robust to the choice of vocabulary. We adopt the expanded vocabulary because it aligns more naturally with the heterogeneous semantics and numeric ranges of different box attributes (see Figure 3 of the main text). Separating the token spaces makes token-type masking and decoding logic more explicit and reduces ambiguity about which parameter a given token represents.

\begin{figure}[t]
\centering
\includegraphics[width=\linewidth]{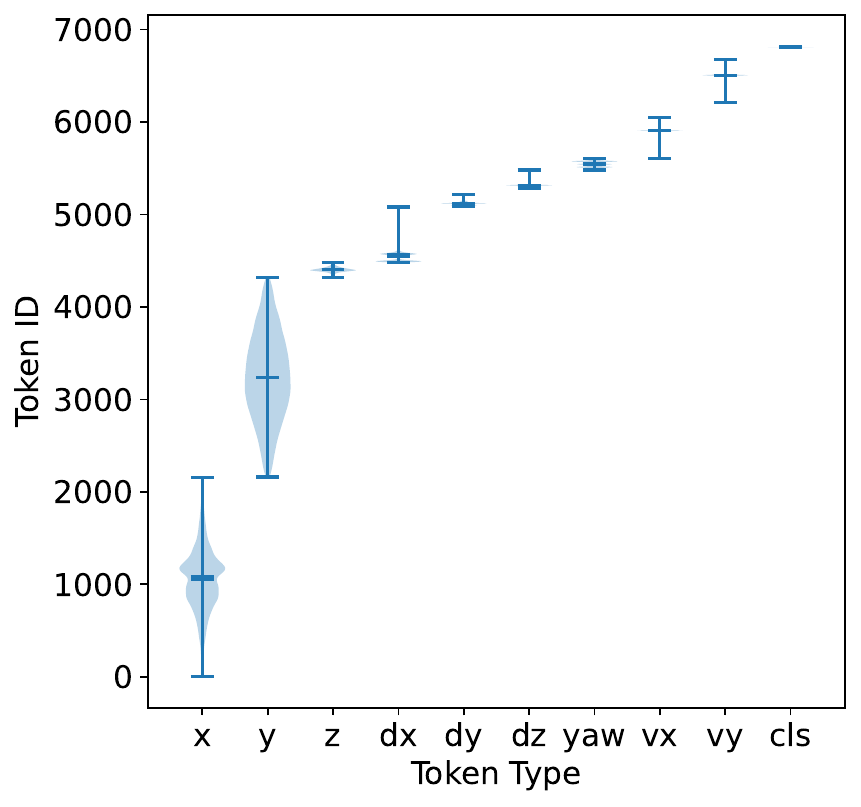}

\vspace{-0.9em}
\caption{\textbf{Token Distribution After Quantization.} Violin plot of the token distribution, plotting token ID by type. Semantic types occupy disjoint ID ranges, with types that cover larger value ranges correspondingly taking up larger token ranges.}
\label{fig:token_distribution}

\vspace{-0.5em}
\end{figure}

\paragraph{Binning Strategies.}
\begin{table}[t]
    \centering
    \small
    \begin{adjustbox}{width=0.8\linewidth, center}
        \begin{tabular}{lccccc}
            \toprule
            Method & Precision & Recall & F1 \\
            \midrule
            Shared                & 74.2 & 59.7 & 65.7 \\
            Expanded                & 74.9 & 59.4 & 65.8 \\
            \bottomrule
        \end{tabular}
    \end{adjustbox}

    \vspace{-0.5em}    \caption{\textbf{Expanded \vs Shared Vocabulary.}}
    \label{tab:supp_vocab}
    \vspace{-2em}
\end{table}
To build our customized vocabulary, we uniformly quantize each continuous box parameter $x, y, z, l, w, h, \psi, v_x, v_y$, into 2160, 2160, 160, 600, 200, 200, 125, 600, 600 bins respectively, which is equivalent to quantize parameters with bin widths $0.05\,\mathrm{m}$ for center/size; $0.05\,\mathrm{rad}$ for yaw; and $0.1\,\mathrm{m/s}$ for velocity. A distribution over the token frequency can be seen in \autoref{fig:token_distribution}.

To justify our choice of token resolution and bin count, we ablate different settings in~\autoref{tab:binning}.  Overall, our method is largely robust to the choice of binning strategy. Reducing the bin count by half (coarser resolution) slightly improves F1, but worsens true-positive (TP) errors; we attribute this to the fact that nuScenes matching is center-distance based and therefore tolerant to small quantization errors, while a smaller vocabulary simplifies prediction, leading to the slight F1 gain. Doubling the bin count (finer resolution) yields lower F1, lower TP errors, but higher memory usage. 
We also attempted adaptive (quantile-based) binning with the same number of bins, which achieves similar F1 to the default setting but worse TP errors, likely due to train-test distribution mismatch. Thus, our default choice strikes a balance between efficiency and performance.

\subsection{RL Decoding Methods}
\begin{table}[t]
    \centering
    \small
    \begin{adjustbox}{width=0.8\linewidth, center}
        \begin{tabular}{lccccc}
            \toprule
            Method & Precision & Recall & F1 \\
            \midrule
            Nucleus     & 72.8 & 60.4 & 65.8 \\
            Greedy      & 74.5 & 60.9 & 66.7 \\
            Beam Search & 74.4 & 60.9 & 66.7 \\
            \bottomrule
        \end{tabular}
    \end{adjustbox}

    \vspace{-0.5em}    
    \caption{\textbf{Analysis on RL Decoding Method.} Different generation decoding strategies have limited effect on performance after RL fine-tuning.}
    \label{tab:rl_decode}
    \vspace{-1em}
\end{table}

We compare the effect of different decoding strategies on our method after RL fine-tuning in \autoref{tab:rl_decode}. Similar to teacher-forcing in \autoref{tab:decode}, nucleus sampling (F1=65.8) performs worse than greedy (F1=66.7) and beam search (F1=66.7). However, unlike teacher-forcing, greedy decoding matches beam search due to RL's sharpening effect on token distributions, causing beams to collapse to similar trajectories, as reported by others~\cite{yue2025RL}. This sharpening enables the RL model to outperform teacher-forcing (F1=61.9) under nucleus sampling.

\section{Additional Details on \ours{}}
\label{sec:supp_imp_details}

\subsection{Additional Implementation Details}
For supervised training, we train all models for 20 epochs with a batch size of 64. We first freeze the encoder and only train the decoder, then we unfreeze and train both the encoder and decoder jointly. We apply a cosine warm-up schedule for the first 10\% of the total training steps. For RL fine-tuning, we train the decoder with a fixed learning rate of $1\times10^{-4}$ and no learning rate scheduling, while keeping the encoder frozen. We use a group size of 8, omit the KL penalty by setting $\beta=0$, and train with a batch size of 64 for around 8 hours. We train all models using automatic mixed precision with BF16. 
To further speed up inference, we adopt KV-caching and an early stopping condition during decoding. Our implementation is based on OpenPCDet\footnote{\url{https://github.com/open-mmlab/OpenPCDet}} and HuggingFace Transformers\footnote{\url{https://github.com/huggingface/transformers}}. All experiments are conducted using 4 NVIDIA A100 (80GB) or 4 H100 GPUs.

\section{Additional Quantitative Results}
\label{sec:supp_quantitative}
We show additional nuScenes quantitative results in \autoref{tab:additional_quantitative}, listing the threshold used for each baseline and TP errors of each method.

\begin{table*}[t]
\centering
\begin{adjustbox}{width=\linewidth}
\begin{tabular}{lccc!{\vrule}ccccc}
    \toprule
    Method & Prec. & Rec. & F1 & mATE~$\downarrow$ & mASE~$\downarrow$ & mAOE~$\downarrow$ & mAVE~$\downarrow$ & mAAE~$\downarrow$ \\
    \midrule
    Ours &  74.9 & 59.4 & 65.8 & 0.323 & 0.278 & 0.362 & 0.234 & 0.191 \\
    Ours~($0.5\times$bin count) & 74.9 & 61.0 & 66.8 & 0.324 & 0.283 & 0.339 & 0.272 & 0.183 \\
    Ours~($2\times$bin count) & 74.3 & 59.9 & 65.8 & 0.316 & 0.276 & 0.357 & 0.218 & 0.182 \\
    \midrule
    Ours~(w/ adaptive binning) & 73.7 & 59.9 & 65.7 & 0.323 & 0.299 & 0.363 & 0.257 & 0.202 \\
    \bottomrule
\end{tabular}
\end{adjustbox}
\caption{\textbf{Binning Strategy Ablation.}}
\label{tab:binning}
\end{table*}

\begin{table*}[t]
\centering
\begin{adjustbox}{width=\linewidth}
\begin{tabular}{lcc!{\vrule}ccc!{\vrule}ccccc}
    \toprule
    Method & Encoder &Thresh. & Prec. & Rec. & F1 & mATE~$\downarrow$ & mASE~$\downarrow$ & mAOE~$\downarrow$ & mAVE~$\downarrow$ & mAAE~$\downarrow$ \\
    \midrule
    PointPillars & \multirow{5}{*}{Pillar Conv.}  & - & 3.4 & 85.8 & 6.2 & 0.353 & 0.263 & 0.346 & 0.290 & 0.198 \\
    CenterPoint &  & - & 5.6 & 82.4 & 10.1 & 0.316 & 0.262 & 0.437 & 0.241 & 0.190 \\
    PointPillars &  & 0.29 & 58.3 & 50.0 & 53.1 & 0.312 & 0.252 & 0.263 & 0.282 & 0.212 \\
    CenterPoint &   & 0.50 & 67.9 & 53.3 & 59.5 & 0.277 & 0.251 & 0.381 & 0.229 & 0.203 \\
    Ours &   & - & 69.6 & 52.4 & 59.2 & 0.365 & 0.285 & 0.424 & 0.262 & 0.195 \\
    \midrule
    SECOND & \multirow{5}{*}{Voxel Conv.} & - & 3.8 & 85.9 & 6.9 & 0.325 & 0.259 & 0.286 & 0.268 & 0.198 \\
    CenterPoint &   & - & 6.8 & 85.3 & 12.1 & 0.292 & 0.255 & 0.381 & 0.217 & 0.181 \\
    SECOND &   & 0.29 & 63.5 & 55.6 & 59.1 & 0.289 & 0.248 & 0.219 & 0.254 & 0.214 \\
    CenterPoint &   & 0.49 & 72.8 & 60.3 & 65.8 & 0.263 & 0.247 & 0.335 & 0.207 & 0.190 \\
    Ours &   & - & 74.9 & 59.4 & 65.8 & 0.323 & 0.278 & 0.362 & 0.234 & 0.191 \\
    \midrule
    DSVT & \multirow{3}{*}{Transformer} & - & 17.6 & 83.3 & 27.4 & 0.270 & 0.248 & 0.276 & 0.228 & 0.189 \\
    DSVT &  & 0.27 & 79.1 & 66.3 & 71.6 & 0.253 & 0.242 & 0.243 & 0.223 & 0.198 \\
     Ours &  & - & 77.0 & 64.1 & 69.5 & 0.295 & 0.281 & 0.304 & 0.238 & 0.194 \\
    \midrule
    LION & \multirow{3}{*}{Mamba} & - & 16.7 & 84.9 & 26.3 & 0.265 & 0.244 & 0.271 & 0.231 & 0.187 \\
    LION &  & 0.27 & 78.6 & 68.3 & 72.5 & 0.247 & 0.239 &  0.234 & 0.228 & 0.193 \\
     Ours &  & - & 77.5 & 65.2 & 70.4 & 0.301 & 0.278 & 0.283 & 0.220 & 0.181 \\
    \bottomrule
\end{tabular}
\end{adjustbox}
\caption{\textbf{Detailed Evaluation Result.}}
\label{tab:additional_quantitative}
\end{table*}

\section{Additional Qualitative Results}
\label{sec:supp_qualitative}
\begin{figure*}[ht]
    \centering
    \includegraphics[width=\textwidth]{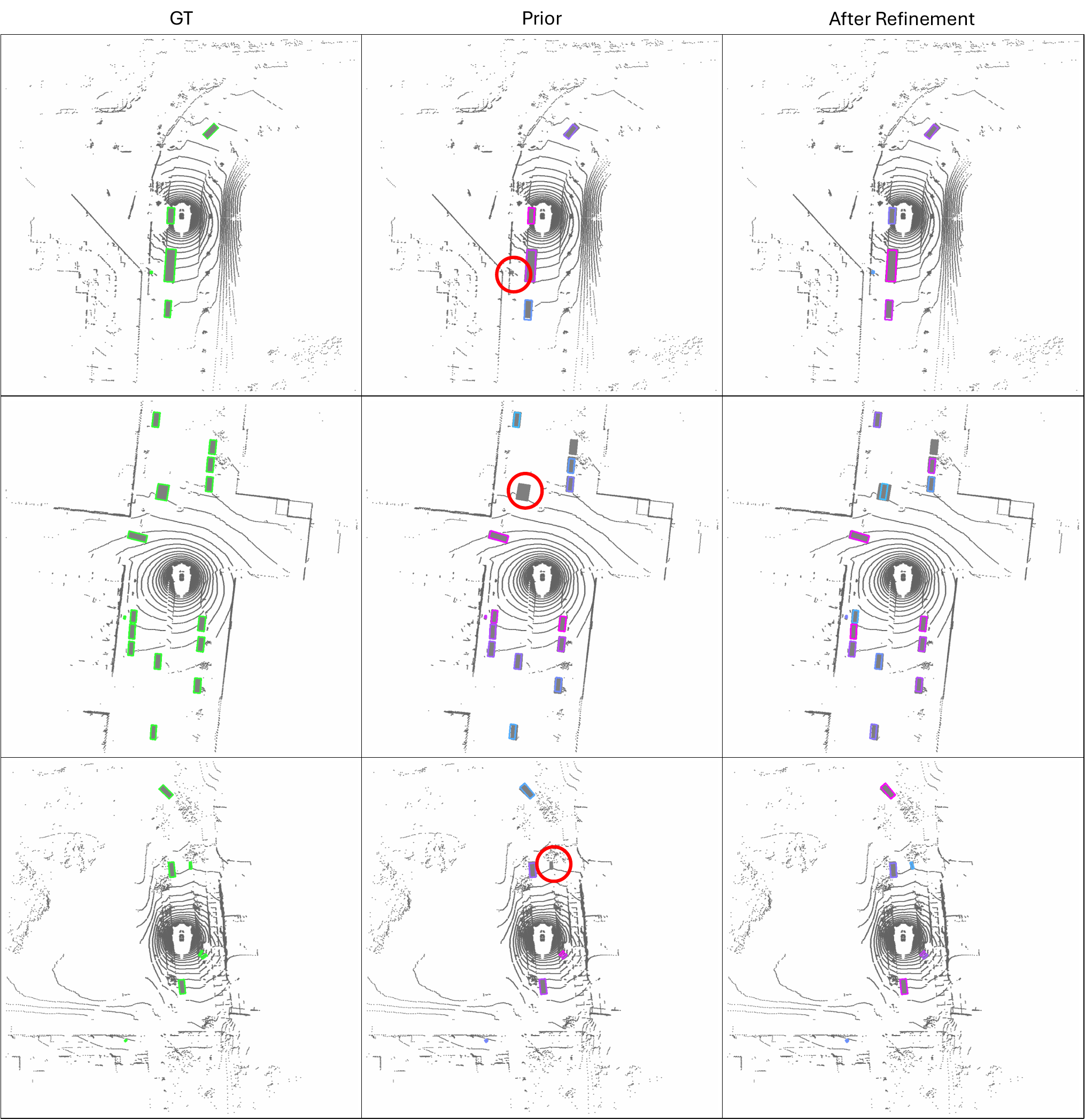}
    \vspace{-0.5em}    \caption{\textbf{Qualitative Results on Cascading Refinement.} 
    We visualize the ground-truth boxes (left), predictions from the near-to-far prior model (middle), and predictions after Cascading Refinement (right). Cascading Refinement recovers objects that were missed by the prior model (circled in \textcolor{red}{red}). Predicted boxes are colored by generation order from first (\textcolor{mymagenta}{magenta}) to last (\textcolor{myblue}{blue}). Best viewed in color.}
    
    \vspace{-0.5em}
    \label{fig:supp_cascade_1}
\end{figure*}

\begin{figure*}[ht]
    \centering
    \includegraphics[width=\textwidth]{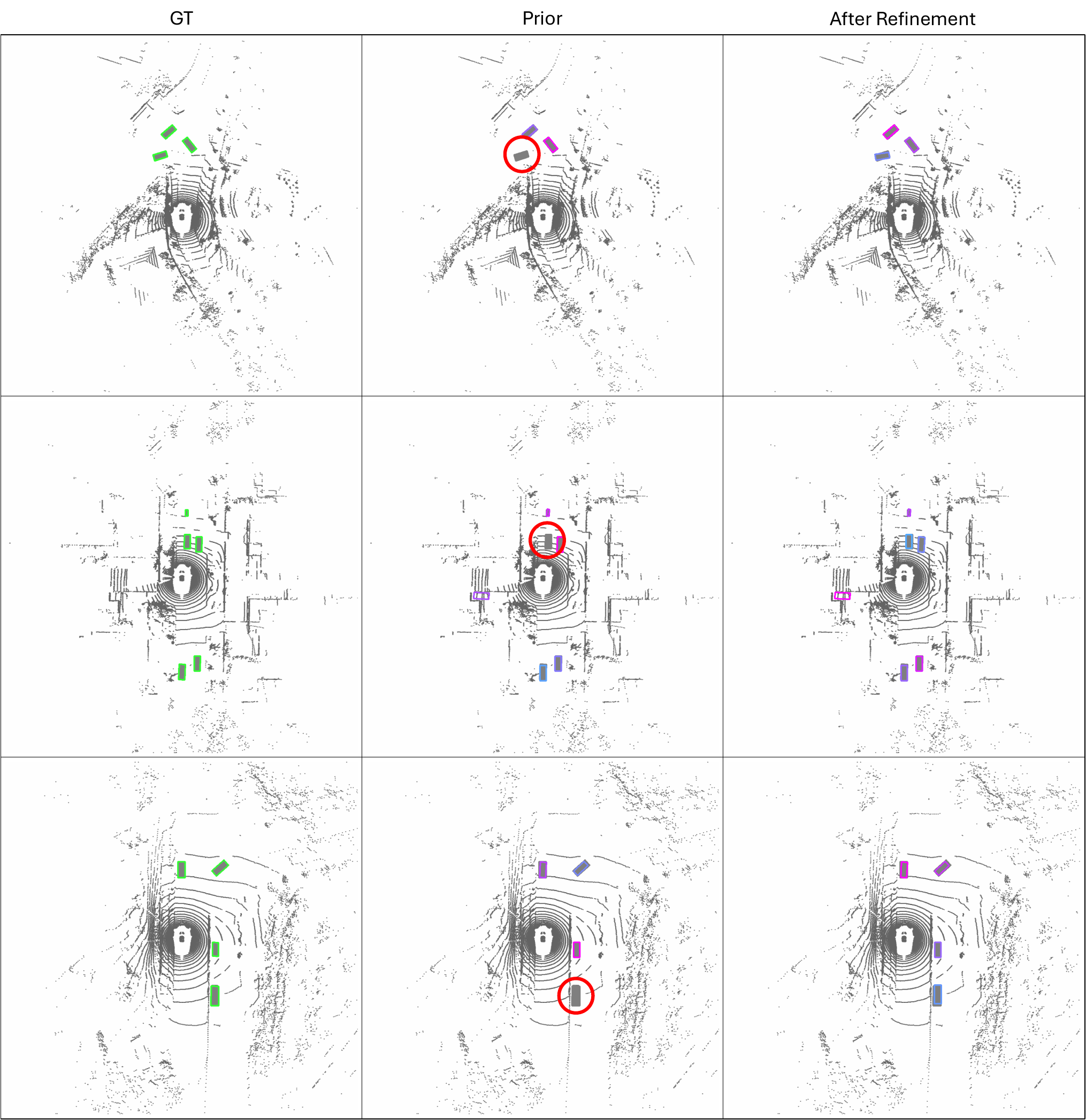}
    \vspace{-0.5em}    \caption{\textbf{Qualitative Results on Cascading Refinement.} 
    We visualize the ground-truth boxes (left), predictions from the near-to-far prior model (middle), and predictions after Cascading Refinement (right). Cascading Refinement recovers objects that were missed by the prior model (circled in \textcolor{red}{red}). Predicted boxes are colored by generation order from first (\textcolor{mymagenta}{magenta}) to last (\textcolor{myblue}{blue}). Best viewed in color.}
    
    \vspace{-0.5em}
    \label{fig:supp_cascade_2}
\end{figure*}

\begin{figure*}[t]
    \centering
    \includegraphics[width=\textwidth]{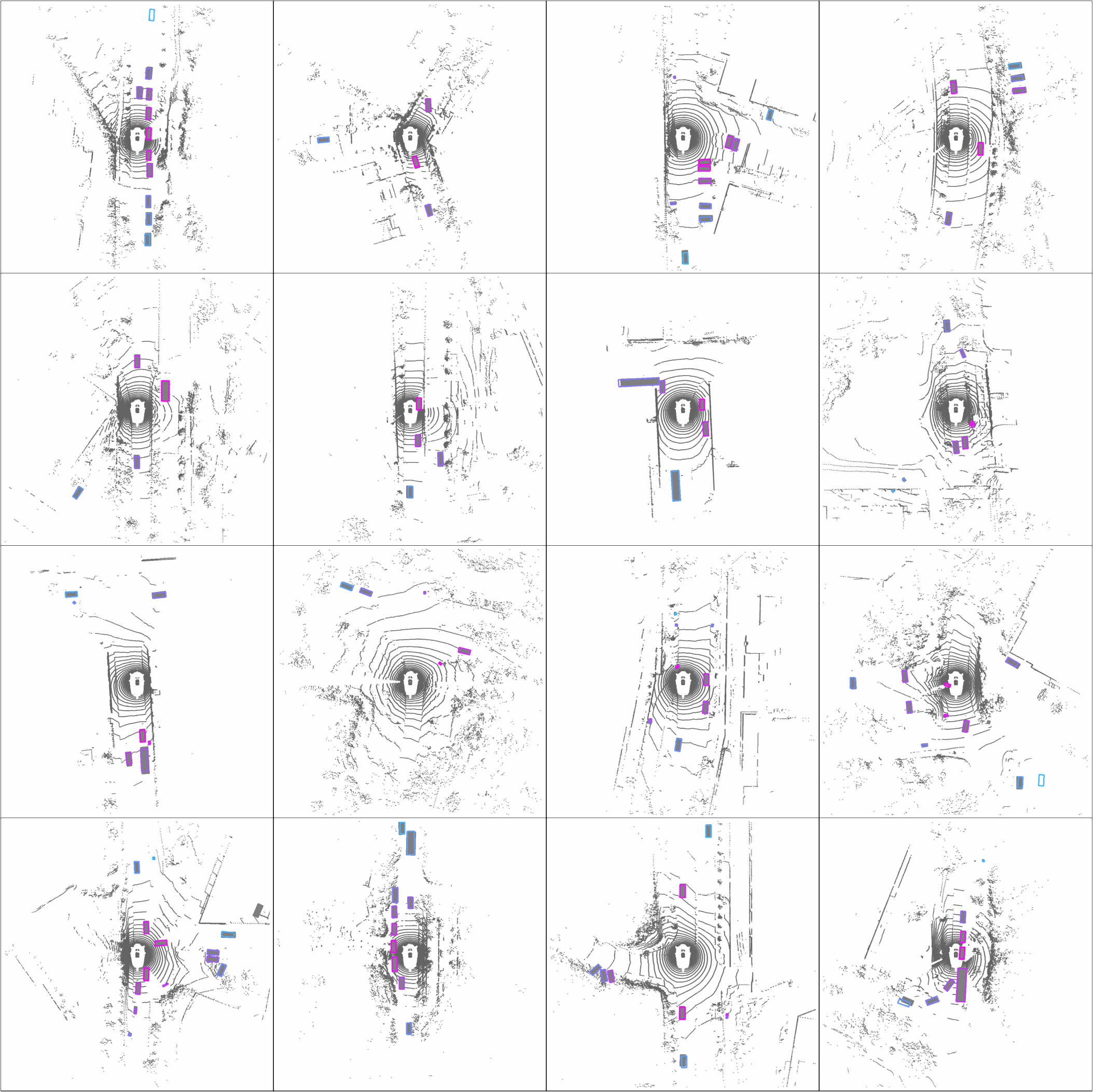}
    \vspace{-0.5em}    \caption{\textbf{Qualitative Results on Pillar-based Backbone.} We visualize bounding box generations from our method with pillar-based backbone trained to generate boxes in near-to-far order. \ours{} generates boxes from first (\textcolor{mymagenta}{magenta}) to last (\textcolor{myblue}{blue}), ground-truth boxes are in gray. Best viewed in color.}
    \vspace{-0.5em}
    \label{fig:supp_pillar_distance}
\end{figure*}

\begin{figure*}[ht]
    \centering
    \includegraphics[width=\textwidth]{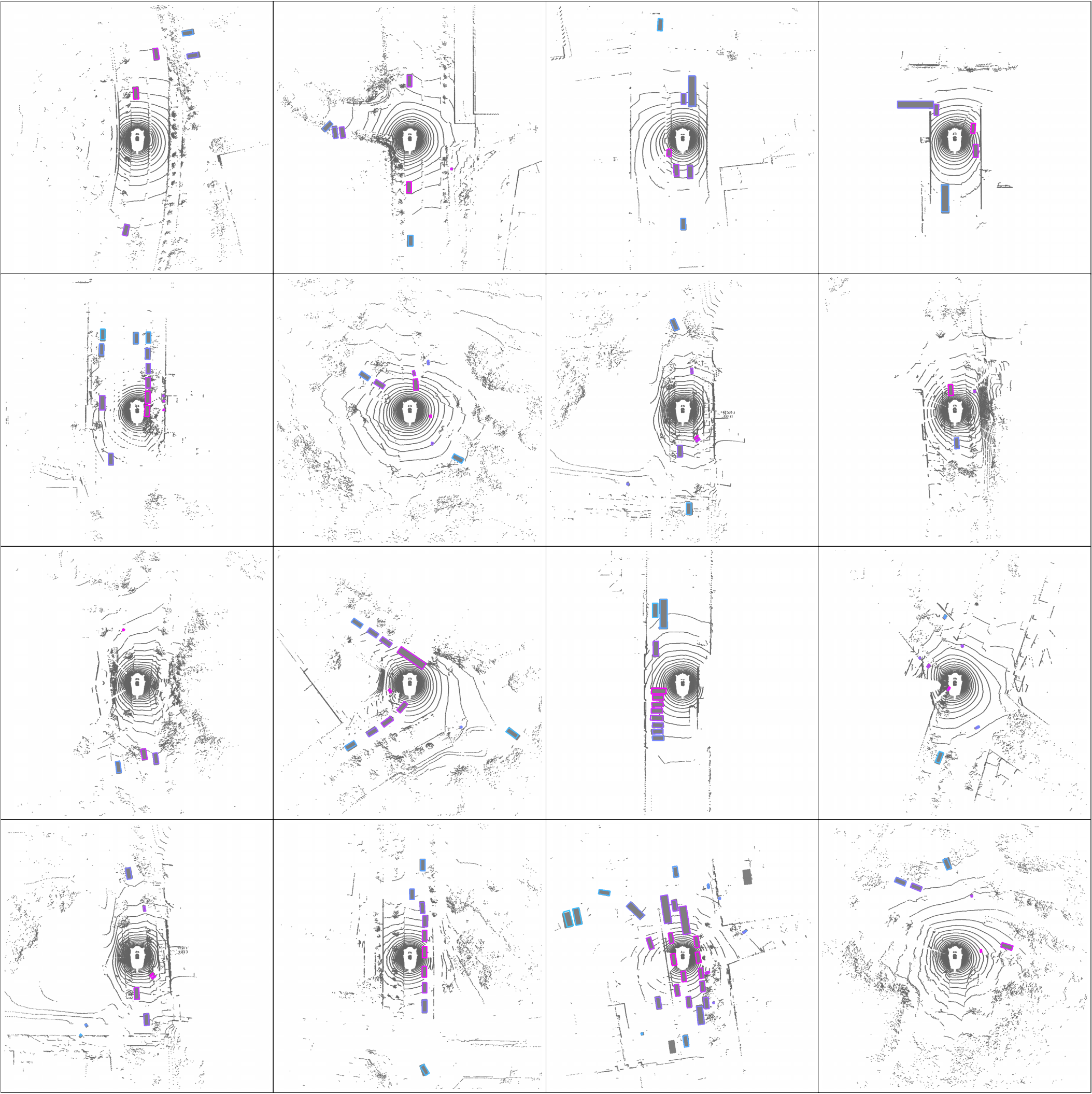}
    \vspace{-0.5em}    \caption{\textbf{Qualitative Results on Voxel-based Backbone.} We visualize bounding box generations from our method with voxel-based backbone trained to generate boxes in near-to-far order. \ours{} generates boxes from first (\textcolor{mymagenta}{magenta}) to last (\textcolor{myblue}{blue}), ground-truth boxes are in gray. Best viewed in color.}
    \vspace{-0.5em}
    \label{fig:supp_voxel_distance}
\end{figure*}

\begin{figure*}[ht]
    \centering
    \includegraphics[width=\textwidth]{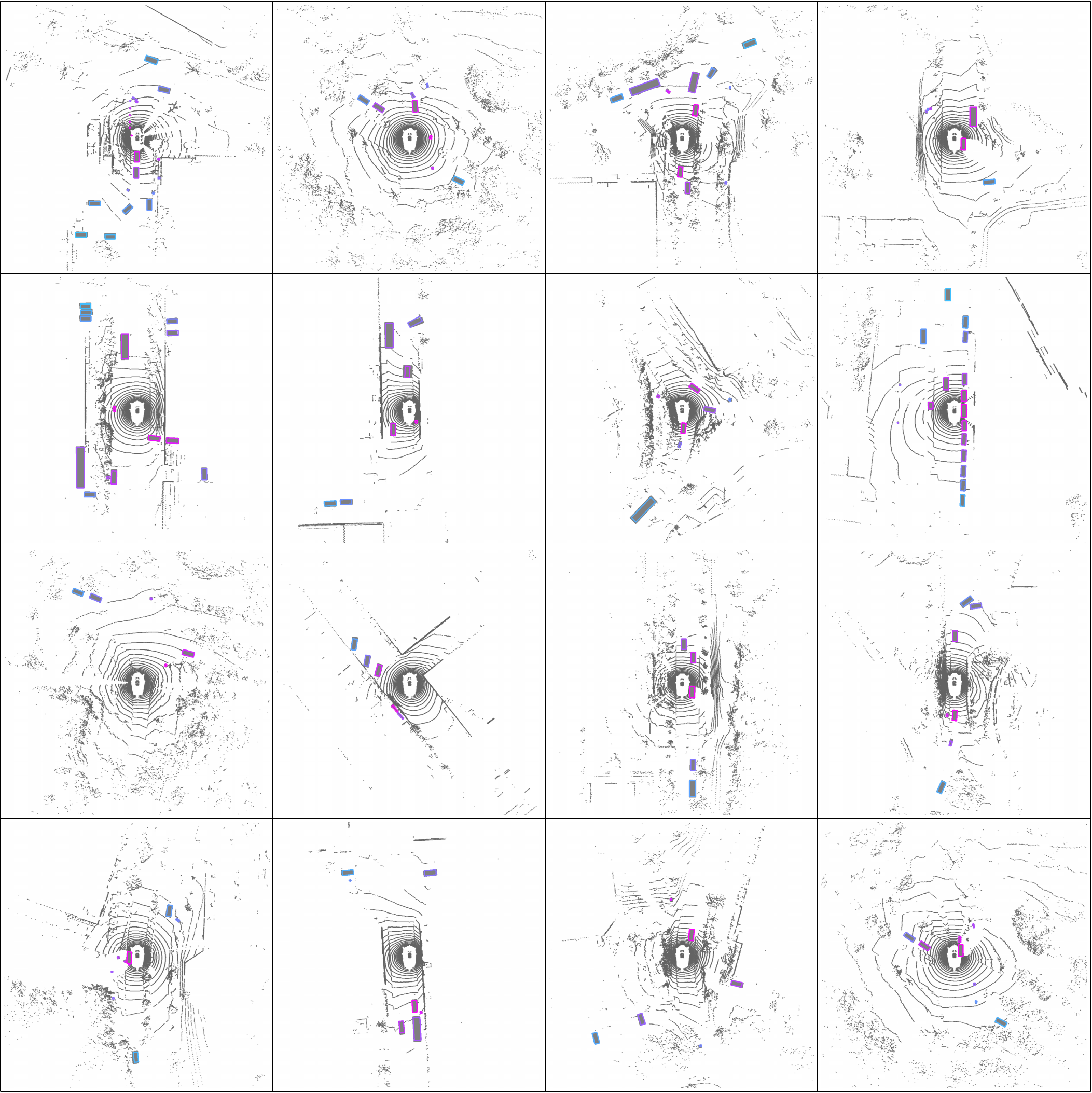}
    \vspace{-0.5em}    \caption{\textbf{Qualitative Results on Transformer-based Backbone.} We visualize bounding box generations from our method with transformer-based backbone trained to generate boxes in near-to-far order. \ours{} generates boxes from first (\textcolor{mymagenta}{magenta}) to last (\textcolor{myblue}{blue}), ground-truth boxes are in gray. Best viewed in color.}
    \vspace{-0.5em}
    \label{fig:supp_dsvt_distance}
\end{figure*}

\begin{figure*}[ht]
    \centering
    \includegraphics[width=\textwidth]{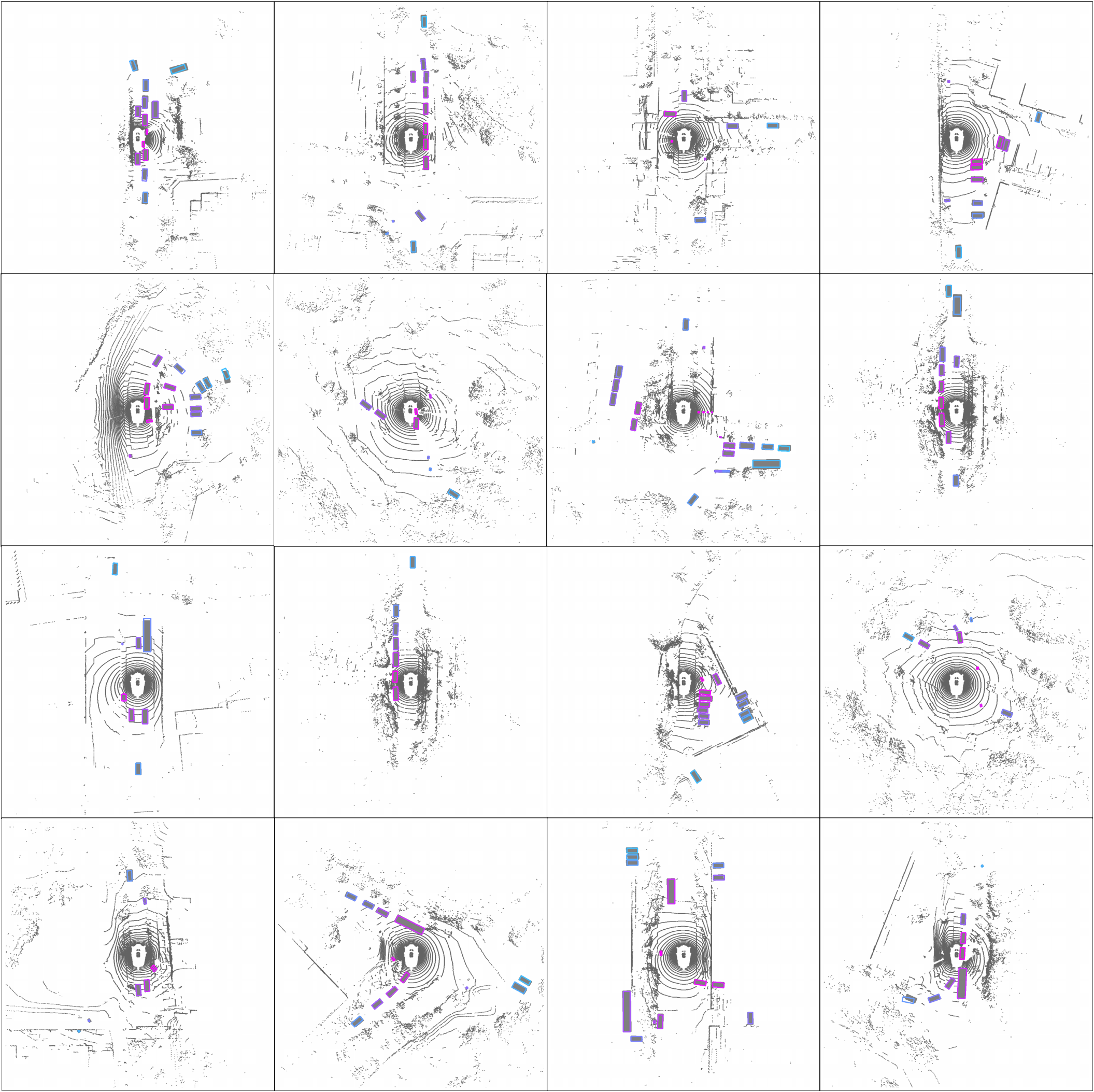}
    \vspace{-0.5em}    \caption{\textbf{Qualitative Results on Mamba-based Backbone.} We visualize bounding box generations from our method with Mamba-based backbone trained to generate boxes in near-to-far order. \ours{} generates boxes from first (\textcolor{mymagenta}{magenta}) to last (\textcolor{myblue}{blue}), ground-truth boxes are in gray. Best viewed in color.}
    \vspace{-0.5em}
    \label{fig:supp_mamba_distance}
\end{figure*}

\begin{figure*}[ht]
    \centering
    \includegraphics[width=\textwidth]{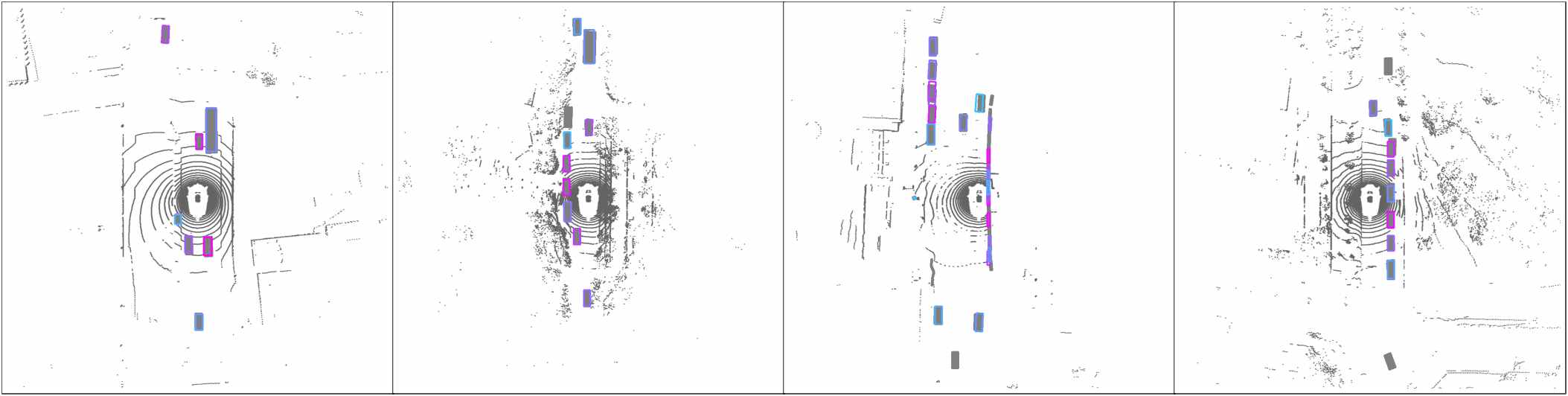}
    \vspace{-0.5em}    \caption{\textbf{Qualitative Results on Random-order Model.} We visualize bounding box generations from our method with voxel-based backbone trained to generate boxes in random order. \ours{} generates boxes from first (\textcolor{mymagenta}{magenta}) to last (\textcolor{myblue}{blue}), ground-truth boxes are in gray. Best viewed in color.}
    
    \vspace{-0.5em}
    \label{fig:supp_voxel_random}
\end{figure*}

\begin{figure*}[ht]
    \centering
    \includegraphics[width=\textwidth]{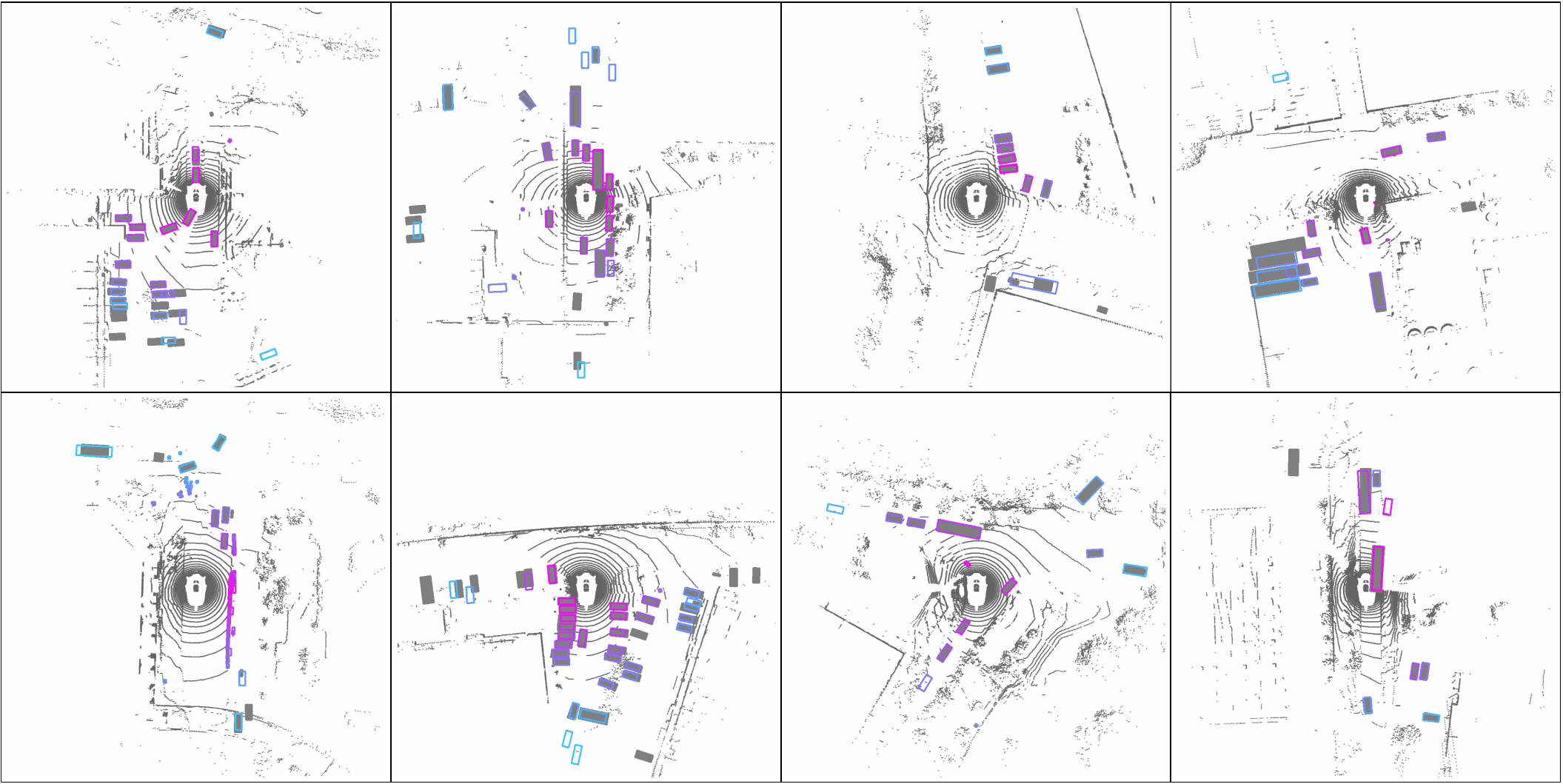}
    \vspace{-0.5em}    \caption{\textbf{Failure Case of \ours{}.} The dominant failure mode of our method include missing detections, extra detections, and misclassifications at distant locations or for sparse objects. \ours{} generates boxes from first (\textcolor{mymagenta}{magenta}) to last (\textcolor{myblue}{blue}), ground-truth boxes are in gray. Best viewed in color.}
    
    \vspace{-0.5em}
    \label{fig:supp_failure_qualitative}
\end{figure*}

We include additional qualitative results:

\begin{itemize}
    \item \autoref{fig:supp_cascade_1} \& \autoref{fig:supp_cascade_2}: Qualitative results on Cascading Refinement.
    \item \autoref{fig:supp_pillar_distance}: Qualitative results on pillar-based backbone.
    \item \autoref{fig:supp_voxel_distance}: Qualitative results on voxel-based backbone.
    \item \autoref{fig:supp_dsvt_distance}: Qualitative results on transformer-based backbone.
    \item \autoref{fig:supp_mamba_distance}: Qualitative results on Mamba-based backbone.
    \item \autoref{fig:supp_voxel_random}: Qualitative results on random-order model.
    \item \autoref{fig:supp_failure_qualitative}: Qualitative examples of \ours{} failure modes.
\end{itemize}

\end{document}